\documentclass{article}

\PassOptionsToPackage{numbers, compress}{natbib}
\usepackage[dblblindworkshop, final]{neurips_2026}
\workshoptitle{Sim2Science: ML with Imperfect Scientific Models}

\usepackage[utf8]{inputenc} %
\usepackage[T1]{fontenc}    %
\usepackage{hyperref}       %
\usepackage{url}            %
\usepackage{booktabs, multirow, makecell}       %
\usepackage{amsfonts, amsmath}       %
\usepackage{nicefrac}       %
\usepackage{microtype}      %
\usepackage[table]{xcolor}  %
\usepackage{graphicx}
\usepackage{float}
\usepackage{wrapfig}
\definecolor{fgtred}{HTML}{C0392B}
\newcommand{\dg}[2]{\cellcolor{fgtred!#1}#2}
\title{Frozen in Time: How a Physics Foundation Model Forgets Under Sim-to-Real Adaptation}

\author{\normalfont%
  \textbf{Haaris Mian}\textsuperscript{1,2,$*$} \qquad
  \textbf{Miles Cranmer}\textsuperscript{1,2} \\[2pt]
  \textbf{Alessandro Favero}\textsuperscript{1,2,$*\dagger$} \qquad
  \textbf{Payel Mukhopadhyay}\textsuperscript{1,2,$*\dagger$} \\[6pt]
  \textsuperscript{1}University of Cambridge \qquad
  \textsuperscript{2}Polymathic AI
}

\begin{document}

\maketitle
{\renewcommand{\thefootnote}{}%
\footnotetext{$^*$Core technical contributors. $^\dagger$Equal advising. Correspondence: \texttt{ham51@cam.ac.uk}}}

\begin{abstract}
  Scientific foundation models (SciFMs) have achieved remarkable performance on 
  a wide range of PDEs, including the ability to efficiently adapt to unseen data distributions with additional finetuning. As with all neural architectures, finetuning such models comes with the risk of \textit{catastrophic forgetting}, whereby adapting to new tasks degrades prior knowledge of the model. 
  We study catastrophic forgetting in Walrus-1.3B, a SciFM for continuum dynamics, over the course of a two-stage sim-to-real adaptation: first finetuning on simulated flow around a cylinder and then adapting to experimental data of the same system. We find that finetuning on the simulations rapidly degrades rollout performance on the original pretraining datasets, with one-step errors increasing up to 35$\times$ after just 20 epochs. Subsequent adaptation to real data further erodes performance on the simulated task. We characterize the potential drivers of forgetting, such as dataset dependence and task representation similarity.  We then show that replaying as little as 1\% simulated data during real-data finetuning recovers most of the lost simulation performance while retaining effective adaptation to the experimental dynamics.
  Our results identify catastrophic forgetting as a practical obstacle to sim-to-real adaptation of SciFMs and show that lightweight replay strategies significantly improve retention.\looseness=-1
\end{abstract}

\section{Introduction}
\label{sec:intro}
In recent years, large-scale scientific foundation models (SciFMs), such as Walrus \citep{mccabe2026walrus}, DPOT-L \citep{hao2024dpot}, Poseidon \citep{NEURIPS2024_84e1b1ec}, and many others \citep{cao2026vicon, holzschuh2025pde, liu2024prosefdmultimodalpdefoundation, liu2026tadpoleautoencodersfoundationmodels, morel2025discolearningdiscoverevolution, NEURIPS2023_e1579096, NEURIPS2024_bc75fa98, NEURIPS2024_d7cb9db5, nguyen2025physix, PhysRevE.111.035304, rautela2026morphpdefoundationmodels, serrano2026testtimegeneralizationphysicsneural}, have adapted transformer-based architectures similar to those used in large language and vision models to the problem of physics simulation, pretraining with vast amounts of numerically simulated PDE data \citep{jhtdb15, yu2023climsim, ohana2024the, NEURIPS2022_0a974713}. This purely data-driven approach to surrogate modeling has seen considerable success across a wide variety of physics domains, including fluid dynamics, weather, and astrophysics \citep{parker2025aion, stachenfeld2022learned, doi:10.1126/science.adi2336}.

To adapt foundation models to downstream tasks or unseen data distributions, additional finetuning is generally required from the pretrained baseline. However, the gains from finetuning are not without tradeoffs. In particular, supervised fine-tuning runs the risk of \textit{catastrophic forgetting}, \citep{8107520, doi:10.1073/pnas.1611835114, 11151751} well-characterized in the vision and language model literature, whereby weight updates made to the model during finetuning result in severe degradation of the pretrained model's prior capabilities. This phenomenon has spurred the growth of a subfield dubbed Continual Learning (CL), aimed at developing methods to continually adapt models to new tasks while preserving prior knowledge and mitigating the problem of catastrophic forgetting \citep{9349197}.

To deploy SciFMs in real-world scientific or engineering applications, models must ultimately make predictions on experimental data. However, a large gap exists between the simulated PDE data used for pretraining and the real, noisy dynamics observed in a fluids laboratory, known as the \textit{sim-to-real gap} \citep{hu2026realpdebench, 9308468}. Recent work demonstrated zero-shot transfer from such models to laboratory data, and additionally showed that a lightweight second finetuning stage on few samples can further improve the results \citep{mukhopadhyay2026emergent}. While finetuning provides a general strategy to adapt a simulation-trained model to experimental fluid flows, it is currently unclear how such adaptation affects the prior knowledge acquired during pretraining, or whether learning from real data further degrades recently-gained simulation knowledge.\looseness=-1

In this work, we study these questions in Walrus-1.3B using a two-stage adaptation setting: we first finetune the pretrained model on simulated flow around a cylinder, and then adapt it to pair experimental particle-image-velocimetry (PIV) data of the same system. We ask
how and what does a large PDE model forget upon exposure to unseen physics or experimental data? What is the severity of forgetting and its onset? Is forgotten physics recoverable or retainable through CL methods such as replay? 
We find that forgetting begins rapidly during simulated-cylinder finetuning, with prediction error on some pretraining datasets increasing by up to $35 \times$ after just 20 epochs. Later adaptation to real PIV data further degrades performance on the simulated task. Remarkably, we show that replaying even a small fraction of simulated data substantially restores the lost performance.

\begin{figure*}[t]
  \centering
  \includegraphics[width=\textwidth]{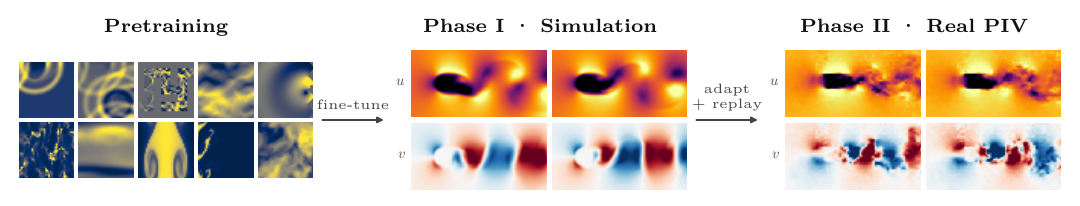}
  \caption{\textbf{Setup.} Walrus-1.3B, pretrained on The Well, is fine-tuned
    on simulated cylinder flow (Phase~I), then adapted to real PIV of the same
    flow (Phase~II). Phase~I induces catastrophic forgetting, so Phase~II
    replays simulation at a fixed fraction.}
  \label{fig:overview}
\end{figure*}

\section{Sim-to-Real Adaptation}
\label{sec:s2r}

We study the adaptation of Walrus \citep{mccabe2026walrus}, a physics foundation model for continuum dynamics, to paired fluid dynamics datasets of flow around a fixed cylinder in two phases.\footnote{See Appendix \ref{sec:exp} for detailed experimental setup and motivation of sim-pretraining phase.} Phase I consists of finetuning the pretrained model on simulated flows, and Phase II consists of adaptation of the Phase I model to real cylinder flows obtained from imaging. We then perform an analysis of forgetting through inference and evaluation on prior data (both from the Well and from cylinder simulation) and attempt to mitigate forgetting in Phase II through a replay-based strategy.\footnote{Code: \url{https://github.com/haarisamian/walrus-forgetting-paper}}

\paragraph{Model.}

Walrus is a 1.3B parameter large physics model trained on 19 different physical scenarios from the Well collection of PDE datasets \citep{ohana2024the}, including acoustic scattering, shear flow, and a wide range of other continuum dynamics datasets. Walrus is an autoregressive transformer model and was pretrained on a next-frame prediction task. In particular, it was trained to output a delta (difference) between the last input frame and the next frame.\looseness=-1

\paragraph{Normalization.}

An important design choice in Walrus is the normalization strategy, applied to a given channel $c$ in the input fields before the forward pass to the model, i.e., $\hat{x}_c = x_c / \sigma_c$ where $\sigma_c = \sqrt{\mathbb{E}[x_c^2]}$, Analogously, $\sigma_c^{\Delta} = \sqrt{\mathbb{E}[\Delta_c^2]}$ is applied to the delta target outputted by the model. This RMS normalization ensures that the model ``sees'' inputs from datasets with widely varying amplitudes at a roughly similar scale, stabilizing prediction and training.\looseness=-1

However, these RMS statistics can be computed in different ways. Under \textit{samplewise normalization} (SW), the model receives an input context which is normalized by its own RMS statistics for each step during training and at inference time. In \textit{global normalization} (GN), the model sees data normalized by RMS statistics across the whole dataset, which preserves amplitude variation between trajectories. We analyze the impact of normalization choice in our experiments, as it strongly affects both adaptation and the dynamics of forgetting.

\begin{figure}[t]
    \centering
    \includegraphics[width=\textwidth]{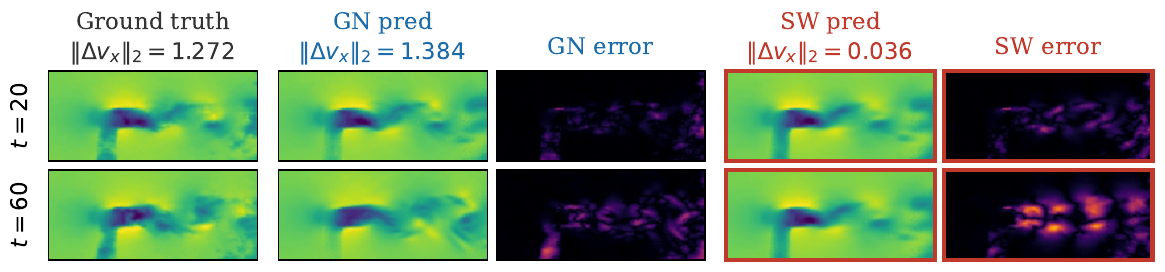}
    \caption{Predicted $v_x$ at rollout steps $t{=}20$ and $t{=}60$ after 40 epochs of Phase~II PIV fine-tuning. \textbf{GN} predictions track the downstream vortex shedding in the ground truth; \textbf{SW} predictions collapse to a near-static field, with $t{=}60$ nearly indistinguishable from $t{=}20$. Both models share training data, epoch count, and rollout trajectory.}
    \label{fig:sw_vs_gn_rollout}
\end{figure}
\paragraph{Cylinder Flow Dataset.}

We finetune the model to the datasets described here, which are previously unseen as vortex shedding or cylinder data is not present in the pretraining data. Crucially, the pretraining data contains no real or experimentally obtained data, so adapting to PIV (Particle Image Velocimetry) presents an even greater challenge for the model. We obtain data from the RealPDEBench paper \citep{hu2026realpdebench}, which provided paired trajectory data for five fluid dynamics scenarios.

The cylinder dataset explores a canonical setup in fluid dynamics, the flow around a rigid cylindrical object, through a set of 92 paired real and simulated trajectories of 3990 steps indexed by Reynolds numbers ranging from 1800–12000. The characteristic phenomenon is von K\'arm\'an vortex shedding, whereby incoming flow leads to an alternating periodic wake of vortices downstream of the cylinder \citep{annurev:/content/journals/10.1146/annurev.fl.28.010196.002401}. The shedding frequency is characteristic of the experimental set-up and scales roughly with the Reynolds number \citep{Roshko1954}. At low Re, the flow is nearly laminar and vortices are regular and smooth, while at increasing Re, the flow becomes more turbulent, and the shedding becomes chaotic while still staying periodic. While simulated data is obtained via numerical simulation in 2D with Lilypad solvers \citep{weymouth2015lilypad}, the real data is obtained through PIV, in which tracer particles are imaged at regular intervals to infer displacement and calculate velocity flows over time.\looseness=-1

\paragraph{Phase I: Simulated Cylinder Finetuning.}

In this phase, we begin from the pretrained checkpoint of Walrus and finetune on cylinder simulation using a range of training protocols, varying both the normalization strategy (samplewise vs. global) and learning rate, and evaluate performance on the new data using one-step and rollout validation, as well as qualitative video rollouts. In addition to this main sweep, we ran a separate, dedicated finetuning trial (global normalization, $\text{lr}=5\times10^{-5}$, 20 epochs, checkpoints every 5 epochs) to study forgetting dynamics at finer granularity in the earliest epochs of finetuning, evaluating each of the early checkpoints against the original Well pretraining datasets.\looseness=-1

\paragraph{Phase II: Real Cylinder Finetuning.}

In Phase II, we continue adapting the Phase I finetuned checkpoint to real PIV cylinder data. Global norm finetuning was the only strategy that successfully learned the real PIV rollout dynamics; samplewise training failed to learn the PIV vortex shedding at all, so we adopt global norm as the backbone for all subsequent Phase II experiments.\looseness=-1

\paragraph{Samplewise normalization can cause frozen rollouts.}
Figure~\ref{fig:sw_vs_gn_rollout} illustrates how SW fine-tuning fails in the real data adaptation phase.
After 40 epochs of PIV training, the SW model's rollout predictions collapse to a
near-static field.
Between rollout steps $t{=}30$ and $t{=}60$ the SW predicted $v_x$ field barely changes, with no visible vortex shedding, while the ground truth continues to evolve
dynamically in line with ground truth flows.
We hypothesize this is caused by the samplewise statistics decaying over the course of rollout on out-of-distribution
real data. The persistent zero-delta prediction learned under samplewise finetuning is a shortcut that prevents Walrus from predicting the core vortex dynamics of the real cylinder setup, even when exposed to simulated cylinder flows.\looseness=-1

\paragraph{Normalization strategy determines Phase II success.}
A central finding is that replacing Walrus's native samplewise normalization with global normalization (GN) before real-data finetuning is necessary for effective sim-to-real transfer. With GN, Phase II training on real cylinder PIV converges steadily over 100 epochs, reaching a 60-step rollout VRMSE (variance-scaled RMSE) of 0.318 on the validation set
(Figure~\ref{fig:training_and_replay}, left), with the model continuing to improve
through the final epoch. The samplewise (SW) variant fails qualitatively before reaching any comparable
performance.\looseness=-1

\section{Catastrophic Forgetting \& Replay}
\label{sec:cat}
\begin{wrapfigure}{r}{0.5\textwidth} 
  \centering 
  \includegraphics[width=0.95\linewidth]{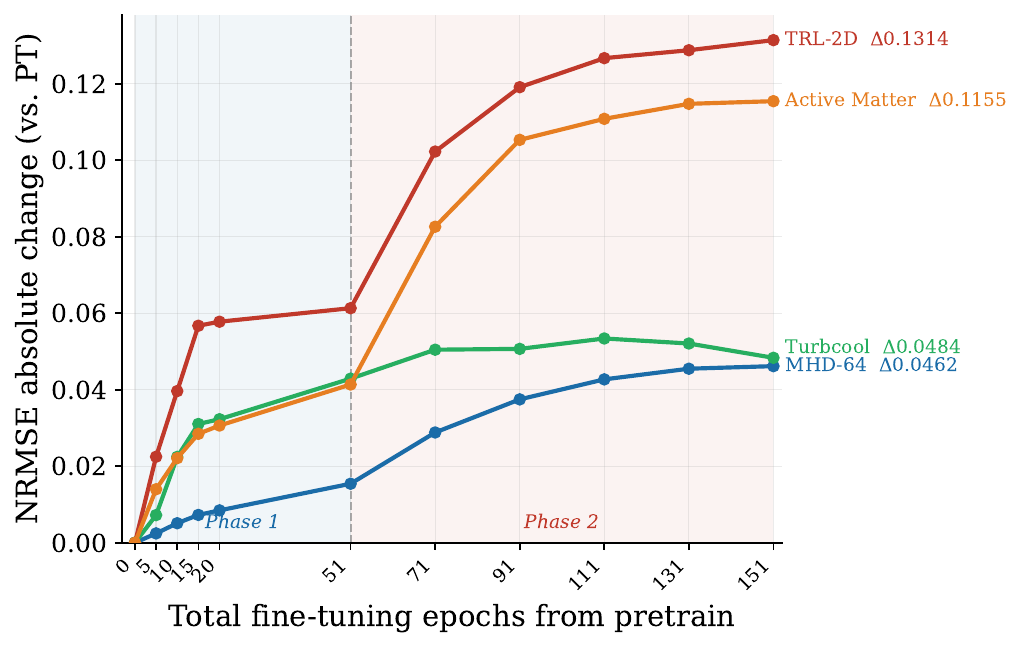} 
  \caption{Forgetting on the Well benchmark across both phases of finetuning. The selected pretraining datasets from the Well degrade mostly monotonically. Active Matter and TRL-D degrade most severely out of these four, while MHD suffers less.} 
  \label{fig:well_forgetting} 
\end{wrapfigure}

\paragraph{Evaluating \& mitigating forgetting.}

To characterize forgetting at fine temporal resolution, we ran a 20-epoch Phase 1 finetuning run (global norm, lr=$5\times10^{-5}$) checkpointed every 5 epochs, and evaluate each checkpoint against ten held-out Well pretraining datasets: MHD, turbulent radiative layer, turbulence gravity cooling, active matter, acoustic scattering, planetary shallow water, Helmholtz staircase, and shear flow. Our primary metric of forgetting in this case is step-1 VRMSE. Additionally, we measure interphase forgetting on the cylinder simulation due to cylinder PIV finetuning. To mitigate the forgetting of Phase I dynamics in Phase II, we explore replaying \citep{NEURIPS2019_fa7cdfad} a fraction of simulated cylinder data alongside real PIV data, sweeping the replay ratio coarsely in increments of 10\% and then finely between 1-10\%.\looseness=-1

\begin{table*}[t]
  \setlength{\belowcaptionskip}{6pt}
  \caption{%
    Catastrophic forgetting during Phase~I cylinder simulation fine-tuning,
    under both error normalizations. NRMSE and VRMSE differ only for fields with
    a non-zero mean. All ratios ($\times$) and absolute increases ($\Delta$) are
    at epoch~20, relative to the pretrained baseline.
    \emph{CKA} is the linear centered kernel alignment between the dataset's and the cylinder
    simulation's activation covariances at the input to the final block of the pretrained
    model. \emph{Pers.} is persistence, the NRMSE of copying the previous frame.
    Both predictors are computed before any fine-tuning.%
  }
  \label{tab:phase1_forgetting}
  \footnotesize
  \setlength{\tabcolsep}{5pt}
  \makebox[\textwidth][c]{%
  \begin{tabular}{lrrr rrr rr}
    \toprule
    & \multicolumn{3}{c}{NRMSE} & \multicolumn{3}{c}{VRMSE}
      & \multicolumn{2}{c}{Predictors} \\
    \cmidrule(lr){2-4} \cmidrule(lr){5-7} \cmidrule(lr){8-9}
    Dataset & Pre & $\times$ & $\Delta$ & Pre & $\times$ & $\Delta$ & CKA & Pers. \\
    \midrule
    \multicolumn{9}{l}{\textit{Fluid dynamics}} \\
    Active Matter    & 0.0106 &  3.88 & \dg{19}{0.0306} & 0.0145 &  4.14 & \dg{23}{0.0455} & 0.027 & 0.312 \\
    TRL-2D           & 0.0380 &  2.52 & \dg{28}{0.0578} & 0.1148 &  2.33 & \dg{41}{0.1525} & 0.146 & 0.250 \\
    Turbcool         & 0.0392 &  1.82 & \dg{20}{0.0323} & 0.0842 &  1.90 & \dg{31}{0.0757} & 0.463 & 0.096 \\
    Shear Flow       & 0.0077 &  2.21 & \dg{1}{0.0093}  & 0.0079 &  2.29 & \dg{0}{0.0102}  & 0.034 & 0.078 \\
    \cmidrule(l){1-9}
    \multicolumn{9}{l}{\textit{Magnetohydrodynamics}} \\
    MHD-64           & 0.1234 &  1.07 & \dg{0}{0.0085}  & 0.1493 &  1.07 & \dg{0}{0.0101}  & 0.015 & 0.356 \\
    \cmidrule(l){1-9}
    \multicolumn{9}{l}{\textit{Wave / acoustics}} \\
    Helmholtz        & 0.0011 & 34.89 & \dg{21}{0.0363} & 0.0013 & 37.66 & \dg{23}{0.0466} & 0.050 & 0.136 \\
    Acoustic (disc.) & 0.0066 & 30.63 & \dg{46}{0.1943} & 0.0072 & 30.80 & \dg{47}{0.2141} & 0.884 & 0.326 \\
    Acoustic (incl.) & 0.0166 & 11.69 & \dg{44}{0.1774} & 0.0178 & 11.80 & \dg{45}{0.1923} & 0.880 & 0.314 \\
    Acoustic (maze)  & 0.0394 & 10.28 & \dg{55}{0.3654} & 0.0398 & 10.30 & \dg{55}{0.3704} & 0.776 & 0.865 \\
    \cmidrule(l){1-9}
    \multicolumn{9}{l}{\textit{Geophysics}} \\
    Planet SWE       & 0.0007 & 28.48 & \dg{12}{0.0193} & 0.0007 & 28.44 & \dg{10}{0.0194} & 0.222 & 0.026 \\
    \bottomrule
  \end{tabular}%
  }
\end{table*}

\paragraph{Catastrophic forgetting is rapid and dataset-dependent.}
Even before real-data fine-tuning begins, Phase~I cylinder simulation fine-tuning
inflicts severe forgetting across the Well benchmark (Figure~\ref{fig:well_forgetting},
Table~\ref{tab:phase1_forgetting}).
The onset of forgetting is remarkably fast. Within the first five epochs of Phase~I fine-tuning, acoustic scattering datasets
exhibit mean NRMSE (normalized RMSE) ratios of $\times$4.6--14.6, and Active
Matter degrades by $\times$2.3 (Figure~\ref{fig:well_forgetting}).
By epoch 20 the wave and geophysics datasets have collapsed catastrophically
(Helmholtz $\times$34.9, Planet SWE $\times$28.5), while MHD barely
moves ($\times$1.07). Absolute degradation ranks the datasets differently: Acoustic (maze) suffers the largest true loss ($\Delta$VRMSE $0.37$), while Helmholtz and Planet SWE post
large ratios only because their pretrained errors were near zero. Either way the spread is wide, a $37\times$ range in
$\Delta$VRMSE.\looseness=-1

\paragraph{Predicting forgetting before finetuning.}
Degradation varies by more than an order of magnitude across the datasets, so we ask whether it can be anticipated before any finetuning. We consider two observables computed on the pretrained model.
We define \emph{persistence} as the one-step NRMSE of simply copying the previous frame, which measures how much a field changes between frames and therefore the temporal evolution the model should capture beyond a trivial copy-forward predictor. 
We then quantify \emph{representational similarity} between each dataset and the cylinder simulation using the linear centered kernel alignment (CKA) \citep{pmlr-v97-kornblith19a}
of their activation covariances, taken at the final block of the model.
Together persistence and CKA account for $82\%$ of the variance in absolute
degradation across the ten datasets. The two appear to capture complementary factors.\footnote{These two predictors are effectively independent
(Variance Inflation Factor $1.03$), and each remains significant after controlling for the other
($p<0.05$).} Higher persistence means more temporal structure is at stake, and so more to lose; higher CKA means a dataset relies on representations closer to those used for the cylinder task, which is consistent with greater interference during fine-tuning. Accordingly, MHD-64 ($\mathrm{CKA}=0.015$) is largely preserved despite its relative high persistence error ($0.356$), whereas the acoustic datasets have substantially higher similarity to the finetuning task ($\mathrm{CKA}\in [0.776,0.884]$), and degrade much more strongly. Within the acoustic datasets, maze has the largest persistence error ($0.865$) and also suffers the greatest absolute degradation of any dataset, consistent with persistence contributing independently of representational similarity.
We refer the reader to Appendix~\ref{sec:exp} for further details.

\begin{figure}[t]
    \centering
    \includegraphics[height=4.4cm]{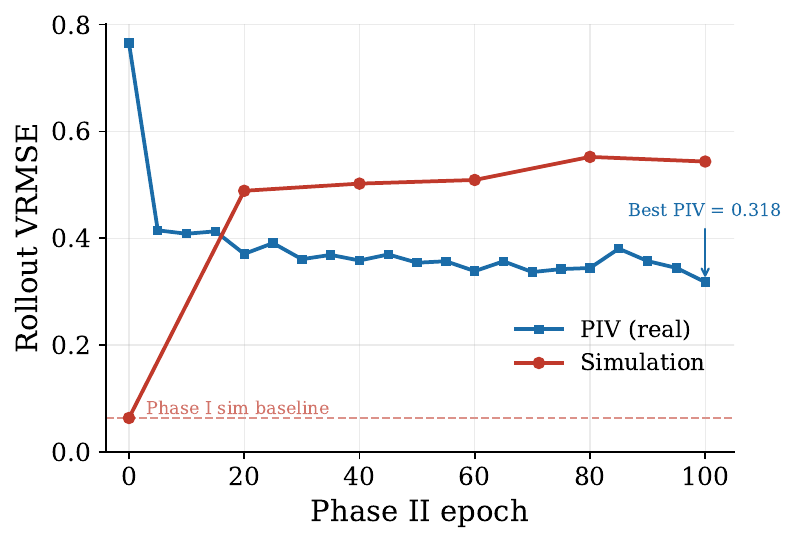}%
    \includegraphics[height=4.4cm]{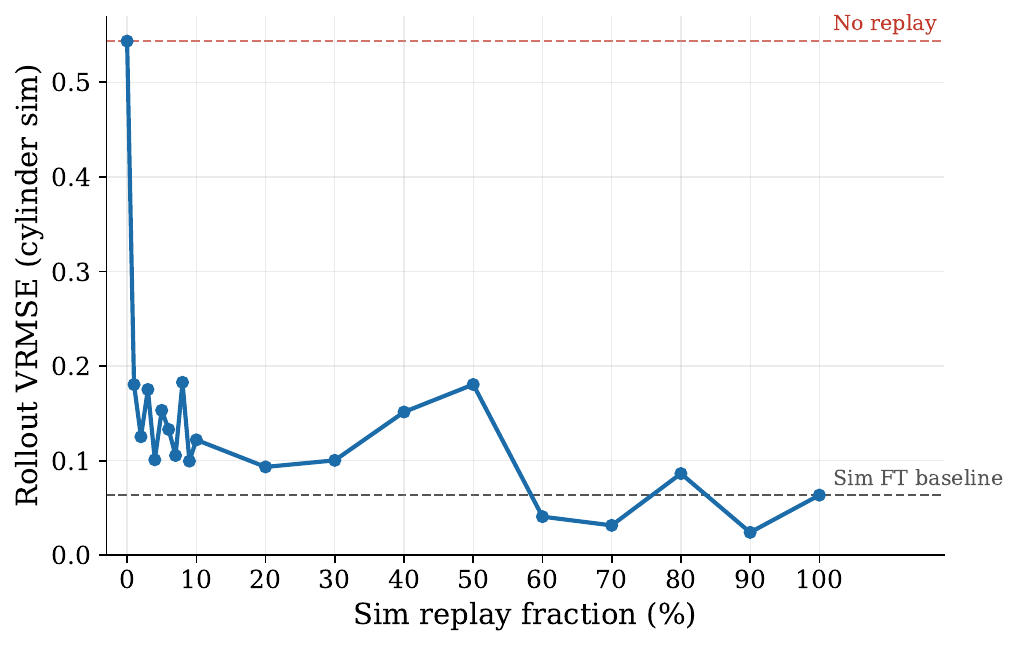}
    \caption{(\textit{Left}) Rollout VRMSE during Phase~II GN training on real
    cylinder PIV reaches 0.318 at epoch 100, while cylinder-simulation skill is
    lost within the first 20 epochs and never recovers. (\textit{Right})
    Cylinder-simulation rollout VRMSE versus the replay fraction mixed into
    Phase~II fine-tuning. Even a 1\% mixture recovers most of the performance
    lost under no-replay training (red dashed).\looseness=-1}
    \label{fig:training_and_replay}
\end{figure}

\paragraph{Sim replay provides an efficient remedy.}
Figure~\ref{fig:training_and_replay} (right) shows that mixing even a small fraction
of Phase I simulation data into Phase II PIV training largely restores cylinder
simulation performance and mitigates catastrophic forgetting on the first task.
Without replay, Phase II GN$\rightarrow$GN training raises rollout NRMSE from
the Phase I baseline of 0.045 to 0.377, an $8\times$ degradation.
At just \textit{1\% simulated replay}, NRMSE falls to below 0.13, recovering $>65\%$ of the lost rollout performance.
Notably, above 60\% replay the model even surpasses the pure
sim finetuning baseline, suggesting that at high fractions the model can continue to improve its performance on cylinder simulations without sacrificing performance on PIV data.\looseness=-1

\section{Conclusion}

In this work, we explored the challenges of adapting large physics models to experimental and out-of-distribution data through fine-tuning and its associated impact on prior task performance. We observe catastrophic forgetting in a scientific foundation model within a few epochs of finetuning, with the severity of forgetting differing by finetuning phase (sim vs. real), persistence, and representational task similarity. Replay mitigates this forgetting on cylinder flow, preserving simulated performance while adapting to experimental data. This controlled characterization opens the door to continual learning methods for large physics models in out-of-distribution settings.

\begin{ack}
Polymathic AI is funded by the Simons Foundation and Schmidt Sciences. This work was supported in part by the AI2050 program at Schmidt Sciences (Grant G-25-70028); M.C. is supported by an AI2050 Early Career Fellowship and the Isaac Newton Trust. The authors thank the Infosys--Cambridge AI Centre and the Swirles HPC system at the University of Cambridge.
\end{ack}

\bibliographystyle{unsrtnat}
\bibliography{references}

\begin{thebibliography}{39}
\providecommand{\natexlab}[1]{#1}
\providecommand{\url}[1]{\texttt{#1}}
\expandafter\ifx\csname urlstyle\endcsname\relax
  \providecommand{\doi}[1]{doi: #1}\else
  \providecommand{\doi}{doi: \begingroup \urlstyle{rm}\Url}\fi

\bibitem[McCabe et~al.(2026)McCabe, Mukhopadhyay, Marwah, Blancard, Rozet, Diaconu, Meyer, Wong, Sotoudeh, Bietti, Morales, Fear, Golkar, Hehir, Hirashima, Krawezik, Lanusse, Morel, Ohana, Parker, Pettee, Shen, Cho, Cranmer, and Ho]{mccabe2026walrus}
Michael McCabe, Payel Mukhopadhyay, Tanya Marwah, Bruno R{\'e}galdo-Saint Blancard, Fran{\c{c}}ois Rozet, Cristiana Diaconu, Lucas~Thibaut Meyer, Kaze W.~K. Wong, Mohammad-Hadi Sotoudeh, Alberto Bietti, Irina~Espejo Morales, Rio~Alexa Fear, Siavash Golkar, Tom Hehir, Keiya Hirashima, Geraud Krawezik, Francois Lanusse, Rudy Morel, Ruben Ohana, Liam~Holden Parker, Mariel Pettee, Jeff Shen, Kyunghyun Cho, Miles Cranmer, and Shirley Ho.
\newblock Walrus: A cross-domain foundation model for continuum dynamics.
\newblock In \emph{Forty-third International Conference on Machine Learning}, 2026.

\bibitem[Hao et~al.(2024)Hao, Su, Liu, Berner, Ying, Su, Anandkumar, Song, and Zhu]{hao2024dpot}
Zhongkai Hao, Chang Su, Songming Liu, Julius Berner, Chengyang Ying, Hang Su, Anima Anandkumar, Jian Song, and Jun Zhu.
\newblock Dpot: Auto-regressive denoising operator transformer for large-scale pde pre-training.
\newblock \emph{arXiv preprint arXiv:2403.03542}, 2024.

\bibitem[Herde et~al.(2024)Herde, Raoni\'{c}, Rohner, K\"{a}ppeli, Molinaro, de~B\'{e}zenac, and Mishra]{NEURIPS2024_84e1b1ec}
Maximilian Herde, Bogdan Raoni\'{c}, Tobias Rohner, Roger K\"{a}ppeli, Roberto Molinaro, Emmanuel de~B\'{e}zenac, and Siddhartha Mishra.
\newblock Poseidon: Efficient foundation models for pdes.
\newblock In A.~Globerson, L.~Mackey, D.~Belgrave, A.~Fan, U.~Paquet, J.~Tomczak, and C.~Zhang, editors, \emph{Advances in Neural Information Processing Systems}, volume~37, pages 72525--72624. Curran Associates, Inc., 2024.
\newblock \doi{10.52202/079017-2311}.
\newblock URL \url{https://proceedings.neurips.cc/paper_files/paper/2024/file/84e1b1ec17bb11c57234e96433022a9a-Paper-Conference.pdf}.

\bibitem[Cao et~al.(2026)Cao, Liu, Yang, Yu, Schaeffer, and Osher]{cao2026vicon}
Yadi Cao, Yuxuan Liu, Liu Yang, Rose Yu, Hayden Schaeffer, and Stanley Osher.
\newblock {VICON}: Vision in-context operator networks for multi-physics fluid dynamics prediction.
\newblock \emph{Transactions on Machine Learning Research}, 2026.
\newblock ISSN 2835-8856.

\bibitem[Holzschuh et~al.(2025)Holzschuh, Liu, Kohl, and Thuerey]{holzschuh2025pde}
Benjamin Holzschuh, Qiang Liu, Georg Kohl, and Nils Thuerey.
\newblock Pde-transformer: Efficient and versatile transformers for physics simulations.
\newblock \emph{arXiv preprint arXiv:2505.24717}, 2025.

\bibitem[Liu et~al.(2024)Liu, Sun, He, Pinney, Zhang, and Schaeffer]{liu2024prosefdmultimodalpdefoundation}
Yuxuan Liu, Jingmin Sun, Xinjie He, Griffin Pinney, Zecheng Zhang, and Hayden Schaeffer.
\newblock Prose-fd: A multimodal pde foundation model for learning multiple operators for forecasting fluid dynamics, 2024.
\newblock URL \url{https://arxiv.org/abs/2409.09811}.

\bibitem[Liu et~al.(2026)Liu, Koehler, Holzschuh, and Thuerey]{liu2026tadpoleautoencodersfoundationmodels}
Qiang Liu, Felix Koehler, Benjamin Holzschuh, and Nils Thuerey.
\newblock Tadpole: Autoencoders as foundation models for 3d pdes with online learning, 2026.
\newblock URL \url{https://arxiv.org/abs/2605.15284}.

\bibitem[Morel et~al.(2025)Morel, Han, and Oyallon]{morel2025discolearningdiscoverevolution}
Rudy Morel, Jiequn Han, and Edouard Oyallon.
\newblock Disco: learning to discover an evolution operator for multi-physics-agnostic prediction, 2025.
\newblock URL \url{https://arxiv.org/abs/2504.19496}.

\bibitem[Subramanian et~al.(2023)Subramanian, Harrington, Keutzer, Bhimji, Morozov, Mahoney, and Gholami]{NEURIPS2023_e1579096}
Shashank Subramanian, Peter Harrington, Kurt Keutzer, Wahid Bhimji, Dmitriy Morozov, Michael Mahoney, and Amir Gholami.
\newblock Towards foundation models for scientific machine learning: Characterizing scaling and transfer behavior.
\newblock In A.~Oh, T.~Naumann, A.~Globerson, K.~Saenko, M.~Hardt, and S.~Levine, editors, \emph{Advances in Neural Information Processing Systems}, volume~36, pages 71242--71262. Curran Associates, Inc., 2023.
\newblock \doi{10.52202/075280-3119}.
\newblock URL \url{https://proceedings.neurips.cc/paper_files/paper/2023/file/e15790966a4a9d85d688635c88ee6d8a-Paper-Conference.pdf}.

\bibitem[Rahman et~al.(2024)Rahman, George, Elleithy, Leibovici, Li, Bonev, White, Berner, Yeh, Kossaifi, Azizzadenesheli, and Anandkumar]{NEURIPS2024_bc75fa98}
Ashiqur Rahman, Robert~Joseph George, Mogab Elleithy, Daniel Leibovici, Zongyi Li, Boris Bonev, Colin White, Julius Berner, Raymond~A. Yeh, Jean Kossaifi, Kamyar Azizzadenesheli, and Anima Anandkumar.
\newblock Pretraining codomain attention neural operators for solving multiphysics pdes.
\newblock In A.~Globerson, L.~Mackey, D.~Belgrave, A.~Fan, U.~Paquet, J.~Tomczak, and C.~Zhang, editors, \emph{Advances in Neural Information Processing Systems}, volume~37, pages 104035--104064. Curran Associates, Inc., 2024.
\newblock \doi{10.52202/079017-3306}.
\newblock URL \url{https://proceedings.neurips.cc/paper_files/paper/2024/file/bc75fa9843a7905bbed9d83895a88f7f-Paper-Conference.pdf}.

\bibitem[McCabe et~al.(2024)McCabe, R\'{e}galdo-Saint~Blancard, Parker, Ohana, Cranmer, Bietti, Eickenberg, Golkar, Krawezik, Lanusse, Pettee, Tesileanu, Cho, and Ho]{NEURIPS2024_d7cb9db5}
Michael McCabe, Bruno R\'{e}galdo-Saint~Blancard, Liam Parker, Ruben Ohana, Miles Cranmer, Alberto Bietti, Michael Eickenberg, Siavash Golkar, Geraud Krawezik, Francois Lanusse, Mariel Pettee, Tiberiu Tesileanu, Kyunghyun Cho, and Shirley Ho.
\newblock Multiple physics pretraining for spatiotemporal surrogate models.
\newblock In A.~Globerson, L.~Mackey, D.~Belgrave, A.~Fan, U.~Paquet, J.~Tomczak, and C.~Zhang, editors, \emph{Advances in Neural Information Processing Systems}, volume~37, pages 119301--119335. Curran Associates, Inc., 2024.
\newblock \doi{10.52202/079017-3791}.
\newblock URL \url{https://proceedings.neurips.cc/paper_files/paper/2024/file/d7cb9db5ade2db7814fbd01ee59f4c7b-Paper-Conference.pdf}.

\bibitem[Nguyen et~al.(2025)Nguyen, Koneru, Li, and Grover]{nguyen2025physix}
Tung Nguyen, Arsh Koneru, Shufan Li, and Aditya Grover.
\newblock Physix: A foundation model for physics simulations.
\newblock \emph{arXiv preprint arXiv:2506.17774}, 2025.

\bibitem[Sun et~al.(2025)Sun, Liu, Zhang, and Schaeffer]{PhysRevE.111.035304}
Jingmin Sun, Yuxuan Liu, Zecheng Zhang, and Hayden Schaeffer.
\newblock Towards a foundation model for partial differential equations: Multioperator learning and extrapolation.
\newblock \emph{Phys. Rev. E}, 111:\penalty0 035304, Mar 2025.
\newblock \doi{10.1103/PhysRevE.111.035304}.
\newblock URL \url{https://link.aps.org/doi/10.1103/PhysRevE.111.035304}.

\bibitem[Rautela et~al.(2026{\natexlab{a}})Rautela, Most, Mansingh, Love, Scheinker, Oyen, Debardeleben, Lawrence, and Biswas]{rautela2026morphpdefoundationmodels}
Mahindra~Singh Rautela, Alexander Most, Siddharth Mansingh, Bradley~C. Love, Alexander Scheinker, Diane Oyen, Nathan Debardeleben, Earl Lawrence, and Ayan Biswas.
\newblock Morph: Pde foundation models with arbitrary data modality, 2026{\natexlab{a}}.
\newblock URL \url{https://arxiv.org/abs/2509.21670}.

\bibitem[Serrano et~al.(2026)Serrano, Han, Oyallon, Ho, and Morel]{serrano2026testtimegeneralizationphysicsneural}
Louis Serrano, Jiequn Han, Edouard Oyallon, Shirley Ho, and Rudy Morel.
\newblock Test-time generalization for physics through neural operator splitting, 2026.
\newblock URL \url{https://arxiv.org/abs/2602.00884}.

\bibitem[Kanov et~al.(2015)Kanov, Burns, Lalescu, and Eyink]{jhtdb15}
Kalin Kanov, Randal Burns, Cristian Lalescu, and Gregory Eyink.
\newblock The johns hopkins turbulence databases: An open simulation laboratory for turbulence research.
\newblock \emph{Computing in Science \& Engineering}, 17\penalty0 (5):\penalty0 10--17, 2015.
\newblock \doi{10.1109/MCSE.2015.103}.

\bibitem[Yu et~al.(2023)Yu, Hannah, Peng, Lin, Bhouri, Gupta, L{\"u}tjens, Will, Behrens, Busecke, Loose, Stern, Beucler, Harrop, Hillman, Jenney, Ferretti, Liu, Anandkumar, Brenowitz, Eyring, Geneva, Gentine, Mandt, Pathak, Subramaniam, Vondrick, Yu, Zanna, Zheng, Abernathey, Ahmed, Bader, Baldi, Barnes, Bretherton, Caldwell, Chuang, Han, HUANG, Iglesias-Suarez, Jantre, Kashinath, Khairoutdinov, Kurth, Lutsko, Ma, Mooers, Neelin, Randall, Shamekh, Taylor, Urban, Yuval, Zhang, and Pritchard]{yu2023climsim}
Sungduk Yu, Walter Hannah, Liran Peng, Jerry Lin, Mohamed~Aziz Bhouri, Ritwik Gupta, Bj{\"o}rn L{\"u}tjens, Justus~Christopher Will, Gunnar Behrens, Julius Busecke, Nora Loose, Charles~I Stern, Tom Beucler, Bryce Harrop, Benjamin~R Hillman, Andrea Jenney, Savannah Ferretti, Nana Liu, Anima Anandkumar, Noah~D Brenowitz, Veronika Eyring, Nicholas Geneva, Pierre Gentine, Stephan Mandt, Jaideep Pathak, Akshay Subramaniam, Carl Vondrick, Rose Yu, Laure Zanna, Tian Zheng, Ryan Abernathey, Fiaz Ahmed, David~C Bader, Pierre Baldi, Elizabeth Barnes, Christopher Bretherton, Peter Caldwell, Wayne Chuang, Yilun Han, YU~HUANG, Fernando Iglesias-Suarez, Sanket Jantre, Karthik Kashinath, Marat Khairoutdinov, Thorsten Kurth, Nicholas Lutsko, Po-Lun Ma, Griffin Mooers, J.~David Neelin, David Randall, Sara Shamekh, Mark~A Taylor, Nathan Urban, Janni Yuval, Guang Zhang, and Michael Pritchard.
\newblock Climsim: A large multi-scale dataset for hybrid physics-{ML} climate emulation.
\newblock In \emph{Thirty-seventh Conference on Neural Information Processing Systems Datasets and Benchmarks Track}, 2023.
\newblock URL \url{https://openreview.net/forum?id=W5If9P1xqO}.

\bibitem[Ohana et~al.(2024)Ohana, McCabe, Meyer, Morel, Agocs, Beneitez, Berger, Burkhart, Dalziel, Fielding, Fortunato, Goldberg, Hirashima, Jiang, Kerswell, Maddu, Miller, Mukhopadhyay, Nixon, Shen, Watteaux, Blancard, Rozet, Parker, Cranmer, and Ho]{ohana2024the}
Ruben Ohana, Michael McCabe, Lucas~Thibaut Meyer, Rudy Morel, Fruzsina~Julia Agocs, Miguel Beneitez, Marsha Berger, Blakesley Burkhart, Stuart~B. Dalziel, Drummond~Buschman Fielding, Daniel Fortunato, Jared~A. Goldberg, Keiya Hirashima, Yan-Fei Jiang, Rich Kerswell, Suryanarayana Maddu, Jonah~M. Miller, Payel Mukhopadhyay, Stefan~S. Nixon, Jeff Shen, Romain Watteaux, Bruno R{\'e}galdo-Saint Blancard, Fran{\c{c}}ois Rozet, Liam~Holden Parker, Miles Cranmer, and Shirley Ho.
\newblock The well: a large-scale collection of diverse physics simulations for machine learning.
\newblock In \emph{The Thirty-eight Conference on Neural Information Processing Systems Datasets and Benchmarks Track}, 2024.
\newblock URL \url{https://openreview.net/forum?id=00Sx577BT3}.

\bibitem[Takamoto et~al.(2022)Takamoto, Praditia, Leiteritz, MacKinlay, Alesiani, Pfl\"{u}ger, and Niepert]{NEURIPS2022_0a974713}
Makoto Takamoto, Timothy Praditia, Raphael Leiteritz, Daniel MacKinlay, Francesco Alesiani, Dirk Pfl\"{u}ger, and Mathias Niepert.
\newblock Pdebench: An extensive benchmark for scientific machine learning.
\newblock In S.~Koyejo, S.~Mohamed, A.~Agarwal, D.~Belgrave, K.~Cho, and A.~Oh, editors, \emph{Advances in Neural Information Processing Systems}, volume~35, pages 1596--1611. Curran Associates, Inc., 2022.
\newblock \doi{10.52202/068431-0117}.
\newblock URL \url{https://proceedings.neurips.cc/paper_files/paper/2022/file/0a9747136d411fb83f0cf81820d44afb-Paper-Datasets_and_Benchmarks.pdf}.

\bibitem[Parker et~al.(2025)Parker, Lanusse, Shen, Liu, Hehir, Sarra, Meyer, Bowles, Wagner-Carena, Qu, Golkar, Bietti, Bourfoune, Cornette, Hirashima, Krawezik, Ohana, Lourie, McCabe, Morel, Mukhopadhyay, Pettee, Cho, Cranmer, and Ho]{parker2025aion}
Liam~Holden Parker, Francois Lanusse, Jeff Shen, Ollie Liu, Tom Hehir, Leopoldo Sarra, Lucas~Thibaut Meyer, Micah Bowles, Sebastian Wagner-Carena, Helen Qu, Siavash Golkar, Alberto Bietti, Hatim Bourfoune, Pierre Cornette, Keiya Hirashima, Geraud Krawezik, Ruben Ohana, Nicholas Lourie, Michael McCabe, Rudy Morel, Payel Mukhopadhyay, Mariel Pettee, Kyunghyun Cho, Miles Cranmer, and Shirley Ho.
\newblock {AION}-1: Omnimodal foundation model for astronomical sciences.
\newblock In \emph{The Thirty-ninth Annual Conference on Neural Information Processing Systems}, 2025.
\newblock URL \url{https://openreview.net/forum?id=6gJ2ZykQ5W}.

\bibitem[Stachenfeld et~al.(2022)Stachenfeld, Fielding, Kochkov, Cranmer, Pfaff, Godwin, Cui, Ho, Battaglia, and Sanchez-Gonzalez]{stachenfeld2022learned}
Kimberly Stachenfeld, Drummond Fielding, Dmitrii Kochkov, Miles Cranmer, Tobias Pfaff, Jonathan Godwin, Can Cui, Shirley Ho, Peter Battaglia, and Alvaro Sanchez-Gonzalez.
\newblock Learned coarse models for efficient turbulence simulation.
\newblock In \emph{International Conference on Learning Representations}, 2022.

\bibitem[Lam et~al.(2023)Lam, Sanchez-Gonzalez, Willson, Wirnsberger, Fortunato, Alet, Ravuri, Ewalds, Eaton-Rosen, Hu, Merose, Hoyer, Holland, Vinyals, Stott, Pritzel, Mohamed, and Battaglia]{doi:10.1126/science.adi2336}
Remi Lam, Alvaro Sanchez-Gonzalez, Matthew Willson, Peter Wirnsberger, Meire Fortunato, Ferran Alet, Suman Ravuri, Timo Ewalds, Zach Eaton-Rosen, Weihua Hu, Alexander Merose, Stephan Hoyer, George Holland, Oriol Vinyals, Jacklynn Stott, Alexander Pritzel, Shakir Mohamed, and Peter Battaglia.
\newblock Learning skillful medium-range global weather forecasting.
\newblock \emph{Science}, 382\penalty0 (6677):\penalty0 1416--1421, 2023.
\newblock \doi{10.1126/science.adi2336}.
\newblock URL \url{https://www.science.org/doi/abs/10.1126/science.adi2336}.

\bibitem[Li and Hoiem(2018)]{8107520}
Zhizhong Li and Derek Hoiem.
\newblock Learning without forgetting.
\newblock \emph{IEEE Transactions on Pattern Analysis and Machine Intelligence}, 40\penalty0 (12):\penalty0 2935--2947, 2018.
\newblock \doi{10.1109/TPAMI.2017.2773081}.

\bibitem[Kirkpatrick et~al.(2017)Kirkpatrick, Pascanu, Rabinowitz, Veness, Desjardins, Rusu, Milan, Quan, Ramalho, Grabska-Barwinska, Hassabis, Clopath, Kumaran, and Hadsell]{doi:10.1073/pnas.1611835114}
James Kirkpatrick, Razvan Pascanu, Neil Rabinowitz, Joel Veness, Guillaume Desjardins, Andrei~A. Rusu, Kieran Milan, John Quan, Tiago Ramalho, Agnieszka Grabska-Barwinska, Demis Hassabis, Claudia Clopath, Dharshan Kumaran, and Raia Hadsell.
\newblock Overcoming catastrophic forgetting in neural networks.
\newblock \emph{Proceedings of the National Academy of Sciences}, 114\penalty0 (13):\penalty0 3521--3526, 2017.
\newblock \doi{10.1073/pnas.1611835114}.
\newblock URL \url{https://www.pnas.org/doi/abs/10.1073/pnas.1611835114}.

\bibitem[Luo et~al.(2025)Luo, Yang, Meng, Li, Zhou, and Zhang]{11151751}
Yun Luo, Zhen Yang, Fandong Meng, Yafu Li, Jie Zhou, and Yue Zhang.
\newblock An empirical study of catastrophic forgetting in large language models during continual fine-tuning.
\newblock \emph{IEEE Transactions on Audio, Speech and Language Processing}, 33:\penalty0 3776--3786, 2025.
\newblock \doi{10.1109/TASLPRO.2025.3606231}.

\bibitem[De~Lange et~al.(2022)De~Lange, Aljundi, Masana, Parisot, Jia, Leonardis, Slabaugh, and Tuytelaars]{9349197}
Matthias De~Lange, Rahaf Aljundi, Marc Masana, Sarah Parisot, Xu~Jia, Ales Leonardis, Gregory Slabaugh, and Tinne Tuytelaars.
\newblock { A Continual Learning Survey: Defying Forgetting in Classification Tasks }.
\newblock \emph{IEEE Transactions on Pattern Analysis \& Machine Intelligence}, 44\penalty0 (07):\penalty0 3366--3385, July 2022.
\newblock ISSN 1939-3539.
\newblock \doi{10.1109/TPAMI.2021.3057446}.
\newblock URL \url{https://doi.ieeecomputersociety.org/10.1109/TPAMI.2021.3057446}.

\bibitem[Hu et~al.(2026)Hu, Feng, Liu, Yan, Deng, Gao, Zheng, Zheng, Yu, Wang, Li, Ma, Zhou, Lu, Fan, and Wu]{hu2026realpdebench}
Peiyan Hu, Haodong Feng, Hongyuan Liu, Tongtong Yan, Wenhao Deng, Tianrun Gao, Rong Zheng, Haoren Zheng, Chenglei Yu, Chuanrui Wang, Kaiwen Li, Zhi-Ming Ma, Dezhi Zhou, Xingcai Lu, Dixia Fan, and Tailin Wu.
\newblock Real{PDEB}ench: A benchmark for complex physical systems with real-world data.
\newblock In \emph{The Fourteenth International Conference on Learning Representations}, 2026.
\newblock URL \url{https://openreview.net/forum?id=y3oHMcoItR}.

\bibitem[Zhao et~al.(2020)Zhao, Queralta, and Westerlund]{9308468}
Wenshuai Zhao, Jorge~Peña Queralta, and Tomi Westerlund.
\newblock Sim-to-real transfer in deep reinforcement learning for robotics: a survey.
\newblock In \emph{2020 IEEE Symposium Series on Computational Intelligence (SSCI)}, pages 737--744, 2020.
\newblock \doi{10.1109/SSCI47803.2020.9308468}.

\bibitem[Mukhopadhyay et~al.(2026)Mukhopadhyay, Nixon, Watteaux, McCabe, Bietti, Cho, Diaconu, Morales, Fouhey, Golkar, et~al.]{mukhopadhyay2026emergent}
Payel Mukhopadhyay, Stefan~S Nixon, Romain Watteaux, Michael McCabe, Alberto Bietti, Kyunghyun Cho, Cristiana Diaconu, Irina~Espejo Morales, David Fouhey, Siavash Golkar, et~al.
\newblock Emergent transfer of a physics foundation model from simulation to laboratory turbulence.
\newblock \emph{arXiv preprint arXiv:2606.01470}, 2026.

\bibitem[Williamson(1996)]{annurev:/content/journals/10.1146/annurev.fl.28.010196.002401}
C~H~K Williamson.
\newblock Vortex dynamics in the cylinder wake.
\newblock \emph{Annual Review of Fluid Mechanics}, 28\penalty0 (Volume 28, 1996):\penalty0 477--539, 1996.
\newblock ISSN 1545-4479.
\newblock \doi{https://doi.org/10.1146/annurev.fl.28.010196.002401}.
\newblock URL \url{https://www.annualreviews.org/content/journals/10.1146/annurev.fl.28.010196.002401}.

\bibitem[Roshko(1954)]{Roshko1954}
Anatol Roshko.
\newblock On the development of turbulent wakes from vortex streets.
\newblock NACA Report 1191, National Advisory Committee for Aeronautics, Washington, D.C., 1954.

\bibitem[Weymouth(2015)]{weymouth2015lilypad}
Gabriel~D. Weymouth.
\newblock Lily pad: Towards real-time interactive computational fluid dynamics.
\newblock \emph{arXiv preprint arXiv:1510.06886}, 2015.
\newblock URL \url{https://arxiv.org/abs/1510.06886}.

\bibitem[Rolnick et~al.(2019)Rolnick, Ahuja, Schwarz, Lillicrap, and Wayne]{NEURIPS2019_fa7cdfad}
David Rolnick, Arun Ahuja, Jonathan Schwarz, Timothy Lillicrap, and Gregory Wayne.
\newblock Experience replay for continual learning.
\newblock In H.~Wallach, H.~Larochelle, A.~Beygelzimer, F.~d\textquotesingle Alch\'{e}-Buc, E.~Fox, and R.~Garnett, editors, \emph{Advances in Neural Information Processing Systems}, volume~32. Curran Associates, Inc., 2019.
\newblock URL \url{https://proceedings.neurips.cc/paper_files/paper/2019/file/fa7cdfad1a5aaf8370ebeda47a1ff1c3-Paper.pdf}.

\bibitem[Kornblith et~al.(2019)Kornblith, Norouzi, Lee, and Hinton]{pmlr-v97-kornblith19a}
Simon Kornblith, Mohammad Norouzi, Honglak Lee, and Geoffrey Hinton.
\newblock Similarity of neural network representations revisited.
\newblock In Kamalika Chaudhuri and Ruslan Salakhutdinov, editors, \emph{Proceedings of the 36th International Conference on Machine Learning}, volume~97 of \emph{Proceedings of Machine Learning Research}, pages 3519--3529. PMLR, 09--15 Jun 2019.
\newblock URL \url{https://proceedings.mlr.press/v97/kornblith19a.html}.

\bibitem[Zhou et~al.(2026)Zhou, Fang, Yousefpour, Wu, Yan, and Narsilio]{physguard}
Changjian Zhou, Junfeng Fang, Negin Yousefpour, Peng Wu, Bin Yan, and Guillermo~A Narsilio.
\newblock Physguard: Fisher-guided gradient projection for sim-to-real neural pde surrogates, 2026.
\newblock URL \url{https://arxiv.org/abs/2606.16602}.

\bibitem[Wang et~al.(2026)Wang, Eshaghi, Zhuang, Rabczuk, and Liu]{wang2026replay}
Yizheng Wang, Mohammad~Sadegh Eshaghi, Xiaoying Zhuang, Timon Rabczuk, and Yinghua Liu.
\newblock Replay-based continual learning for physics-informed neural operators, 2026.
\newblock URL \url{https://arxiv.org/abs/2605.04832}.

\bibitem[Tripura and Chakraborty(2023)]{tripura2023foundational}
Tapas Tripura and Souvik Chakraborty.
\newblock A foundational neural operator that continuously learns without forgetting, 2023.
\newblock URL \url{https://arxiv.org/abs/2310.18885}.

\bibitem[Hu et~al.(2022)Hu, yelong shen, Wallis, Allen-Zhu, Li, Wang, Wang, and Chen]{hu2022lora}
Edward~J Hu, yelong shen, Phillip Wallis, Zeyuan Allen-Zhu, Yuanzhi Li, Shean Wang, Lu~Wang, and Weizhu Chen.
\newblock Lo{RA}: Low-rank adaptation of large language models.
\newblock In \emph{International Conference on Learning Representations}, 2022.
\newblock URL \url{https://openreview.net/forum?id=nZeVKeeFYf9}.

\bibitem[Rautela et~al.(2026{\natexlab{b}})Rautela, Mansingh, Pachalieva, Most, Hickmann, O'Malley, Scheinker, Love, Oyen, DeBardeleben, Lawrence, and Biswas]{rautela2026outofdistribution}
Mahindra~Singh Rautela, Siddharth Mansingh, Aleksandra Pachalieva, Alexander Most, Kyle~S. Hickmann, Daniel O'Malley, Alexander Scheinker, Bradley~C. Love, Diane Oyen, Nathan~A. DeBardeleben, Earl Lawrence, and Ayan Biswas.
\newblock Out-of-distribution transfer of {PDE} foundation models to material dynamics under extreme loading.
\newblock In \emph{AI{\&}PDE: ICLR 2026 Workshop on AI and Partial Differential Equations}, 2026{\natexlab{b}}.
\newblock URL \url{https://openreview.net/forum?id=yRC7DqRYsg}.

\end{thebibliography}

\appendix

\section{Limitations}
\label{sec:limitations}
We characterize forgetting in a single scientific foundation model (Walrus, 1.3B)
adapted to a single physical scenario (cylinder flow), with one seed per
configuration, so degradation values carry no error bars and we cannot say how
far the specific findings transfer. Our predictive model is fitted on ten
datasets with two predictors, and the block at which similarity is measured was
selected from a sweep over those same datasets; validating it properly would
require a fine-tuning target other than cylinder flow, and the same two factors
already explain considerably less variance in Phase~II than in Phase~I.
Forgetting is quantified through pointwise error metrics, which are insensitive
to failure modes that matter physically (e.g., a model may retain low pointwise error
while losing small-scale spectral fidelity, violating conservation laws, or
mispredicting turbulence statistics), so physics-relevant diagnostics would
likely reveal degradation ours does not register. Finally, our replay experiments
interleave cylinder simulation rather than Well data, addressing retention of the
Phase~I task rather than of the full pretraining corpus, which is left to future work.

\section{Further Related Work}
\label{app:rel}

In one related study on Walrus, \citet{mukhopadhyay2026emergent}
fine-tune the model on a handful of direct numerical simulations of the
Rayleigh--Taylor instability and apply it zero-shot to sliding-barrier laboratory
measurements, finding that the model leaves the DNS-like regime and enters the
experimentally observed mixing-growth band without having seen an experimental
sample. Their work asks whether simulation-trained capability transfers to the
laboratory, and we ask additionally what adapting to the laboratory costs in terms of forgetting and characterize the effects on pretraining capability.

Closest in experimental setting, \citet{physguard} study sim-to-real adaptation
on RealPDEBench, including real cylinder flow, and propose Fisher-guided gradient projection to protect
physics-critical parameters. Their operators are trained from scratch on the
matched numerical data (FNO $50.4$M, CNO $8.0$M, DeepONet $3.5$M, Transolver
$4.3$M), so no broad pretraining corpus exists whose loss could be measured.
Degradation is instead assessed as low-frequency spectral preservation on the
target system itself.

Continual learning in scientific machine learning has otherwise been studied on
small operators over sequences of simulated tasks. \citet{wang2026replay} combine
replay, distillation and LoRA on a $0.37$M-parameter Transolver across a ten-task
Darcy sequence, and \citet{tripura2023foundational} freeze a pretrained ensemble of operator blocks
and learn only a small task-specific gate per PDE, storing one gate per task, an
adapter-style scheme in the manner of LoRA \citep{hu2022lora}. Shared weights are
never updated after pretraining, so forgetting is avoided by construction rather
than mitigated; this is unavailable when the goal is a single model that has
genuinely adapted to the target domain, as here. Work on large pretrained PDE
foundation models has focused on forward transfer: \citet{rautela2026outofdistribution} fine-tune
Poseidon and MORPH on OOD material dynamics simulations without evaluating retention.

To our knowledge, no prior work fine-tunes a billion-parameter PDE foundation
model on real experimental measurements and then measures what is lost across its
pretraining corpus, which is the question this paper addresses.

\section{Experimental Setup}
\label{sec:exp}

\begin{table}[h!]
\centering
\caption{\textbf{Phase~I simulation pretraining improves real-PIV rollout at every stage of Phase~II training.} Paired ablation: both arms start from Walrus-1.3B and are trained on real PIV with identical settings ($\mathrm{lr}=5{\times}10^{-5}$, 200 steps/epoch); they differ only in whether the cylinder-simulation Phase~I stage was run first. Epoch~0 is the zero-shot rollout of each starting checkpoint. Metrics are 64-step rollout means on the held-out PIV split. Bold marks the better arm.}
\label{tab:appendix_pretrain_benefit}
\small
\setlength{\tabcolsep}{4.5pt}
\begin{tabular}{rrrcrrcrr}
\toprule
 & \multicolumn{3}{c}{VRMSE $\downarrow$} & \multicolumn{3}{c}{NRMSE $\downarrow$} & \multicolumn{2}{c}{Pearson $r$ $\uparrow$} \\
\cmidrule(lr){2-4}\cmidrule(lr){5-7}\cmidrule(lr){8-9}
PIV epochs & \makecell{no\\Phase~I} & \makecell{+\,Phase~I} & gain & \makecell{no\\Phase~I} & \makecell{+\,Phase~I} & gain & \makecell{no\\Phase~I} & \makecell{+\,Phase~I} \\
\midrule
  0 & 3.902 & \textbf{0.765} & 5.10$\times$ & 1.817 & \textbf{0.447} & 4.06$\times$ & 0.548 & \textbf{0.653} \\
  20 & 0.484 & \textbf{0.371} & 1.31$\times$ & 0.312 & \textbf{0.238} & 1.31$\times$ & 0.862 & \textbf{0.898} \\
  40 & 0.471 & \textbf{0.358} & 1.31$\times$ & 0.317 & \textbf{0.229} & 1.38$\times$ & 0.856 & \textbf{0.903} \\
  60 & 0.517 & \textbf{0.338} & 1.53$\times$ & 0.346 & \textbf{0.211} & 1.64$\times$ & 0.834 & \textbf{0.916} \\
  80 & 0.441 & \textbf{0.344} & 1.28$\times$ & 0.280 & \textbf{0.210} & 1.34$\times$ & 0.869 & \textbf{0.914} \\
  100 & 0.350 & \textbf{0.318} & 1.10$\times$ & 0.219 & \textbf{0.198} & 1.11$\times$ & 0.913 & \textbf{0.924} \\
\bottomrule
\end{tabular}
\end{table}

\begin{table}[t]
\centering
\caption{\textbf{The benefit of Phase~I is concentrated at long rollout horizons.} Rollout VRMSE ($\downarrow$) on real PIV, decomposed into the time windows logged during validation, for the same paired arms as Table~\ref{tab:appendix_pretrain_benefit}. Once Phase~II training has run, the two arms are indistinguishable over the first few steps; the gap that remains is almost entirely in the $T{=}8{:}22$ and $T{=}22{:}64$ windows, indicating that Phase~I buys long-horizon stability rather than one-step accuracy. Epoch~0 is the zero-shot rollout of each starting checkpoint.}
\label{tab:appendix_pretrain_benefit_horizon}
\small
\setlength{\tabcolsep}{4pt}
\begin{tabular}{rlrrrrr}
\toprule
PIV epochs & Arm & $T{=}0{:}1$ & $T{=}1{:}2$ & $T{=}2{:}8$ & $T{=}8{:}22$ & $T{=}22{:}64$ \\
\midrule
  \multirow{2}{*}{0} & no Phase~I & 0.076 & 0.143 & 0.576 & 2.716 & 4.910 \\
   & +\,Phase~I & \textbf{0.054} & \textbf{0.089} & \textbf{0.207} & \textbf{0.478} & \textbf{0.964} \\
\addlinespace[2pt]
  \multirow{2}{*}{20} & no Phase~I & 0.047 & 0.069 & 0.134 & 0.278 & 0.614 \\
   & +\,Phase~I & 0.047 & \textbf{0.068} & \textbf{0.133} & \textbf{0.254} & \textbf{0.453} \\
\addlinespace[2pt]
  \multirow{2}{*}{40} & no Phase~I & 0.047 & 0.068 & 0.134 & 0.280 & 0.595 \\
   & +\,Phase~I & 0.047 & 0.068 & \textbf{0.131} & \textbf{0.251} & \textbf{0.436} \\
\addlinespace[2pt]
  \multirow{2}{*}{60} & no Phase~I & 0.046 & 0.068 & 0.134 & 0.286 & 0.661 \\
   & +\,Phase~I & 0.047 & 0.068 & \textbf{0.131} & \textbf{0.250} & \textbf{0.407} \\
\addlinespace[2pt]
  \multirow{2}{*}{80} & no Phase~I & 0.046 & 0.068 & 0.130 & 0.268 & 0.553 \\
   & +\,Phase~I & 0.046 & 0.068 & 0.133 & \textbf{0.252} & \textbf{0.415} \\
\addlinespace[2pt]
  \multirow{2}{*}{100} & no Phase~I & 0.044 & 0.065 & 0.126 & 0.242 & 0.428 \\
   & +\,Phase~I & 0.044 & 0.065 & \textbf{0.125} & \textbf{0.232} & \textbf{0.383} \\
\bottomrule
\end{tabular}
\end{table}

\subsection{Cylinder Data}
We prepared the data for our experiments by converting the datasets to HDF5 in the Well format, and by downsampling the simulated data (originally 128 x 256) to match the resolution of our real PIV data (64 x 128), so that the sim-to-real gap could be studied independently of resolution differences. We also limited our training and evaluation to the two shared fields between the datasets (x- and y-velocity). The 92 trajectories were divided into train/valid/test splits at an approximate ratio of 80/10/10, with each split containing a range of Reynolds numbers across the dataset.
\subsection{Phase I: Simulation Finetuning}
We used AdamW (weight decay $10^{-4}$) with three learning rates ($2\times10^{-5}$, $1\times10^{-4}$, and $5\times10^{-5}$), crossed with both normalization strategies, for six finetuning runs total, each initialized from the same pretrained checkpoint with an inverse-square-root learning rate schedule (20-epoch linear warmup, 10-epoch cooldown). Runs were trained for up to 51 epochs, with checkpoints saved every 20 epochs and one-step/rollout validation every 5 epochs and every epoch consisting of 200 batched optimizer steps. Based on Phase I results, the $2\times10^{-5}$ condition was not promising and was dropped from Phase II.

\subsection{Phase II: PIV Data Finetuning}
We keep the same architecture, optimizer, and per-epoch validation cadence as Phase I (AdamW, batch size 1, delta prediction, one-step and rollout validation every 5 epochs, 60-step rollouts), but training for up to 100 epochs at lr=$5\times10^{-5}$ with a shortened warmup/cooldown (3/5 epochs, vs. 20/10 in Phase I) appropriate to continuing from an already-adapted checkpoint rather than initializing from scratch. We compare three initialization/normalization conditions: continuing from the Phase I global-norm checkpoint under global norm, continuing from the Phase I samplewise checkpoint but switching to global norm, and continuing from the Phase I samplewise checkpoint and remaining samplewise throughout.
\paragraph{Persistence baseline.}
Persistence is the error of the trivial predictor $y(t+1)=y(t)$: each frame is
predicted by copying its predecessor. We score it with NRMSE computed per
trajectory, timestep and channel over the spatial axes and then averaged, using
the same channel set the model is scored on. Frames are consecutive---the
configuration sets \texttt{min\_dt\_stride}~$=$~\texttt{max\_dt\_stride}~$=1$ for
every dataset, so consecutive stored frames are what the model steps between. We
average over up to 32 validation trajectories per dataset (as few as nine where
the validation split is smaller) and up to 128 frames per trajectory, apply the
same field transforms used in training (base-10 logarithm of density and
temperature for turbulence gravity cooling), and reduce the two three-dimensional
datasets to their central $z$ slice. The baseline uses no model and no training,
and is computed directly from the raw data. As a sanity check, the pretrained
model outperforms it on every dataset, by between $2.5\times$ (turbulence gravity
cooling) and $126\times$ (Helmholtz).

\paragraph{Representational similarity.}
We record activations by instrumenting the forward pass of the evaluation above,
so that data loading, normalization and checkpoint handling are identical. For each dataset we accumulate the first and
second moments of the 1408-dimensional hidden state at the input of transformer
blocks $\{0,5,10,15,20,25,30,39\}$, over 64 batches with up to 4096 randomly
subsampled space-time tokens each ($262{,}144$ tokens per dataset), and form the
centered covariance. Recording the first
moment matters, since at block~0 the mean direction alone carries roughly $63\%$ of the
total energy, and an uncentered second moment is dominated by it. Similarity
between two datasets is the normalized Frobenius inner product of their
covariances,
$\langle C_A,C_B\rangle_F/(\lVert C_A\rVert_F\lVert C_B\rVert_F)$, which is
confined to $[0,1]$ because covariances are positive semi-definite. All
covariances are computed with pretrained weights, before any fine-tuning, and the
reference is the cylinder simulation. We report block~39; by that depth every
dataset occupies the same 1408-dimensional space irrespective of its field count
(two to eleven), resolution, or spatial dimensionality, which is what makes the
comparison well defined.

The encoders themselves do
differ by dimensionality. Since the cylinder data is two-dimensional, finetuning leaves
the 3D encoder and decoder of MHD-64 and turbulence gravity cooling exactly
unchanged, which may potentially contribute to MHD-64's weak forgetting. The effect appears
limited, however, since turbulence gravity cooling degrades substantially under
the same untouched encoder, and the encoders hold only $6.8\%$ of parameters
against $86.4\%$ in the shared processor, suggesting that forgetting propagates
mainly through shared weights, as it must also for Helmholtz, which has no
velocity channel yet shows the largest relative degradation.

\paragraph{Statistical analysis.}
Degradation and both predictors span more than an order of magnitude, and the
proposed mechanism is multiplicative, so we fit ordinary least squares of
$\log_{10}\Delta\mathrm{VRMSE}$ on $\log_{10}$ of the two predictors. With $n=10$
datasets and two predictors we report adjusted $R^{2}$ throughout, together with
the variance inflation factor and the coefficient $t$-tests, which establish that
each predictor contributes after controlling for the other. The fitted model is
$\Delta\mathrm{VRMSE}\propto\mathrm{CKA}^{0.60}\,\mathrm{Pers.}^{0.61}$
($R^{2}=86.3\%$, adjusted $82.4\%$; $F_{2,7}=22.1$, $p<0.001$), with no dataset
mispredicted by more than $2.3\times$ across a $37\times$ range.

\paragraph{Compute.}
All experiments ran on an internal Slurm cluster using NVIDIA L40S and A100
GPUs. Fine-tuning used four GPUs per job under a 12-hour wall-clock limit, with
longer runs chained across successive jobs; evaluation and analysis jobs were
single-GPU. The six-condition Phase~I sweep consumed approximately 290
GPU-hours and the three Phase~II conditions approximately 600. The replay sweep
dominates the budget at roughly 3{,}300 GPU-hours across 150 training jobs,
since each replay fraction requires a full Phase~II run. Evaluation is
comparatively cheap: held-out forgetting evaluations averaged 31 minutes each
(224 evaluations, 97 GPU-hours), replay evaluations a further 88 GPU-hours, and
activation capture 22 minutes each (55 captures, 16 GPU-hours). The persistence
baseline is CPU-only and completes in under 10 minutes for all ten datasets.
In total the experiments reported here consumed approximately 4{,}400
GPU-hours; the broader project, including preliminary runs, ablations and
experiments held for future work, consumed several hundred GPU-hours more.

\section{Additional Forgetting Detail}
\begin{table*}[h]
\centering
\setlength{\belowcaptionskip}{6pt}
\caption{Well benchmark mean per-field error at each Phase~I forgetting-dynamics checkpoint, under both normalizations. Metric: mean of per-field \texttt{\{N,V\}RMSE\_T=all\_mean} from the one-step validation pass. NRMSE divides by $\mathrm{RMS}(y)$, VRMSE by $\sigma(y)$; the two coincide for zero-mean fields and diverge in proportion to a channel's uniform background, so VRMSE is the comparable metric across physically different channels and $\mathrm{VRMSE}=1$ means no better than predicting the spatial mean. Bold = lowest (best) per row. \emph{Pretrained}: Walrus-1.3B before any cylinder finetuning. \emph{Step $k$}: after $k$ epochs of Phase~I cylinder finetuning (200 gradient steps/epoch).}
\label{tab:appendix_well_dyn}
\footnotesize
\setlength{\tabcolsep}{4pt}
\begin{tabular}{lrrrrrrrrrr}
\toprule
 & \multicolumn{5}{c}{NRMSE} & \multicolumn{5}{c}{VRMSE} \\
\cmidrule(lr){2-6} \cmidrule(lr){7-11}
\textbf{Dataset} & \textbf{pretrained} & \textbf{step 5} & \textbf{step 10} & \textbf{step 15} & \textbf{step 20} & \textbf{pretrained} & \textbf{step 5} & \textbf{step 10} & \textbf{step 15} & \textbf{step 20} \\
\midrule
  acoustic\_disc. & \textbf{0.0066} & 0.0956 & 0.1643 & 0.2016 & 0.2008 & \textbf{0.0072} & 0.1060 & 0.1810 & 0.2229 & 0.2213 \\
  acoustic\_incl. & \textbf{0.0166} & 0.0974 & 0.1679 & 0.1961 & 0.1940 & \textbf{0.0178} & 0.1058 & 0.1815 & 0.2121 & 0.2101 \\
  acoustic\_maze & \textbf{0.0394} & 0.1793 & 0.3019 & 0.3895 & 0.4047 & \textbf{0.0398} & 0.1818 & 0.3061 & 0.3950 & 0.4102 \\
  active\_matter & \textbf{0.0106} & 0.0246 & 0.0328 & 0.0391 & 0.0413 & \textbf{0.0145} & 0.0388 & 0.0493 & 0.0571 & 0.0599 \\
  helmholtz & \textbf{0.0011} & 0.0052 & 0.0165 & 0.0304 & 0.0374 & \textbf{0.0013} & 0.0074 & 0.0215 & 0.0383 & 0.0478 \\
  MHD\_64 & \textbf{0.1234} & 0.1258 & 0.1285 & 0.1307 & 0.1318 & \textbf{0.1493} & 0.1522 & 0.1554 & 0.1580 & 0.1594 \\
  planetswe & \textbf{0.0007} & 0.0125 & 0.0178 & 0.0199 & 0.0200 & \textbf{0.0007} & 0.0126 & 0.0179 & 0.0200 & 0.0201 \\
  shear\_flow & \textbf{0.0077} & 0.0122 & 0.0149 & 0.0168 & 0.0170 & \textbf{0.0079} & 0.0129 & 0.0158 & 0.0179 & 0.0181 \\
  TRL\_2D & \textbf{0.0380} & 0.0605 & 0.0777 & 0.0948 & 0.0958 & \textbf{0.1148} & 0.1691 & 0.2179 & 0.2643 & 0.2674 \\
  turbcool & \textbf{0.0392} & 0.0464 & 0.0616 & 0.0702 & 0.0715 & \textbf{0.0842} & 0.1029 & 0.1383 & 0.1572 & 0.1599 \\
\bottomrule
\end{tabular}
\end{table*}

\begin{figure}[H]
  \centering
  \includegraphics[width=\textwidth,height=0.48\textheight,keepaspectratio]{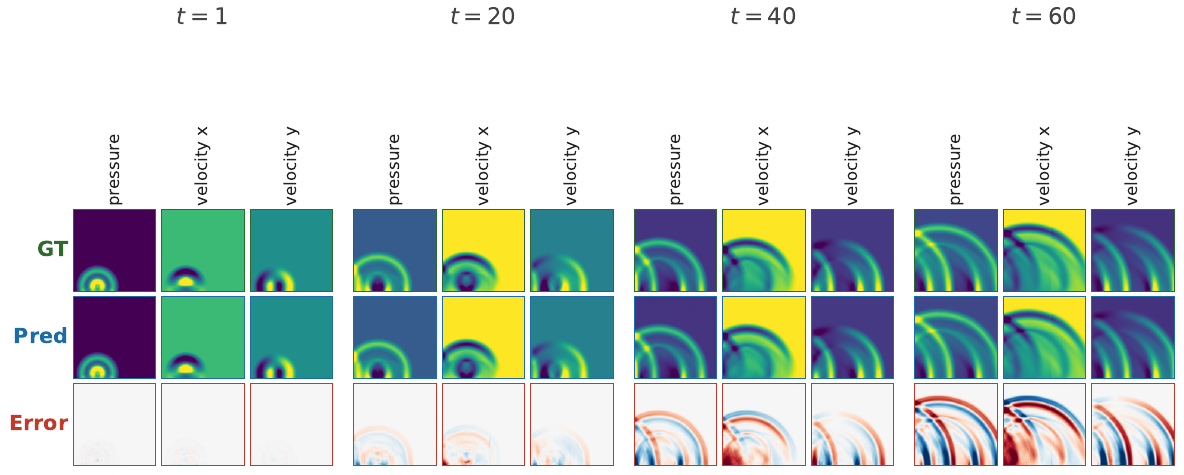}\\[4pt]
  \includegraphics[width=\textwidth,height=0.48\textheight,keepaspectratio]{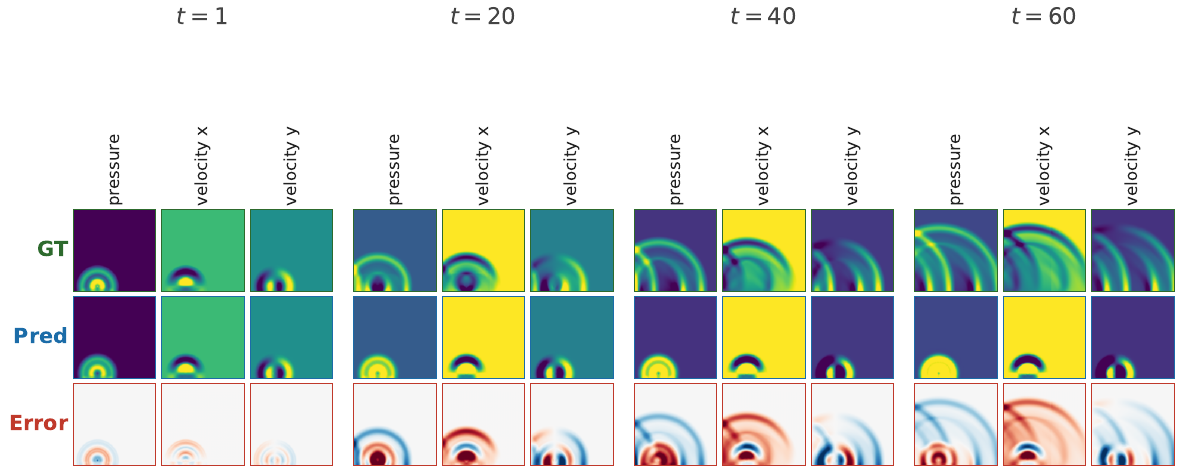}
  \caption{\textbf{Acoustic Scattering (Discontinuous).}
    GT, prediction, and signed error (shared colorscale per field) at rollout steps $t=1,20,40,60$.
    \emph{Top}: pretrained Walrus-1.3B (before Phase~I finetuning).
    \emph{Bottom}: after Phase~I cylinder finetuning (step~20).}
  \label{fig:appendix_well_asd}
\end{figure}

\begin{figure}[H]
  \centering
  \includegraphics[width=\textwidth,height=0.48\textheight,keepaspectratio]{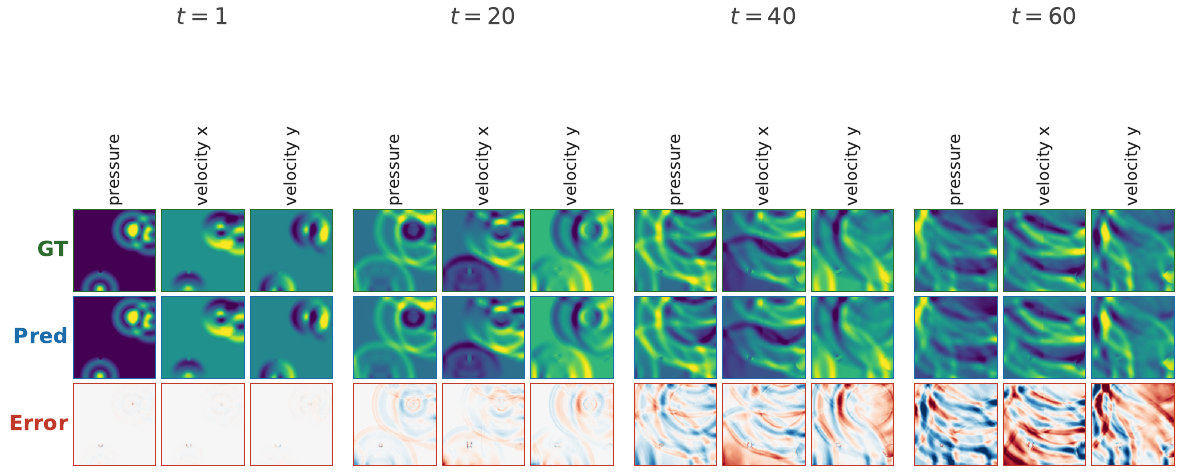}\\[4pt]
  \includegraphics[width=\textwidth,height=0.48\textheight,keepaspectratio]{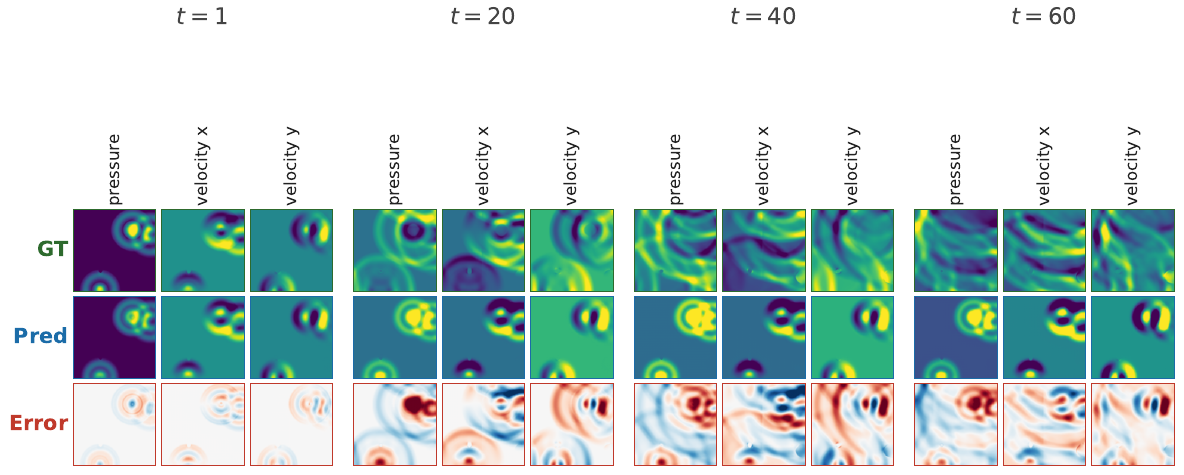}
  \caption{\textbf{Acoustic Scattering (Inclusions).}
    \emph{Top}: pretrained. \emph{Bottom}: Phase~I step~20.}
  \label{fig:appendix_well_asi}
\end{figure}

\begin{figure}[H]
  \centering
  \includegraphics[width=\textwidth,height=0.48\textheight,keepaspectratio]{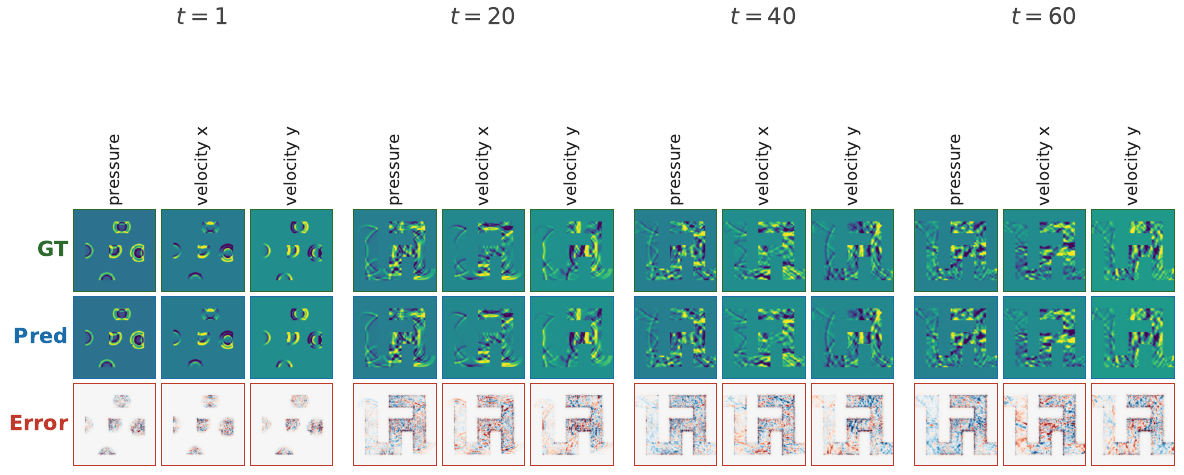}\\[4pt]
  \includegraphics[width=\textwidth,height=0.48\textheight,keepaspectratio]{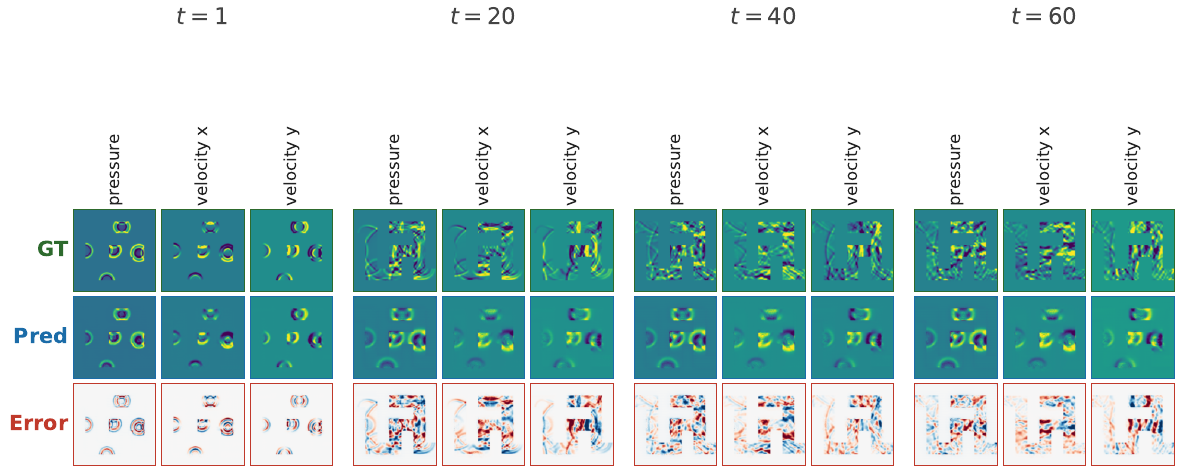}
  \caption{\textbf{Acoustic Scattering (Maze).}
    \emph{Top}: pretrained. \emph{Bottom}: Phase~I step~20.}
  \label{fig:appendix_well_asm}
\end{figure}

\begin{figure}[H]
  \centering
  \includegraphics[width=\textwidth,height=0.48\textheight,keepaspectratio]{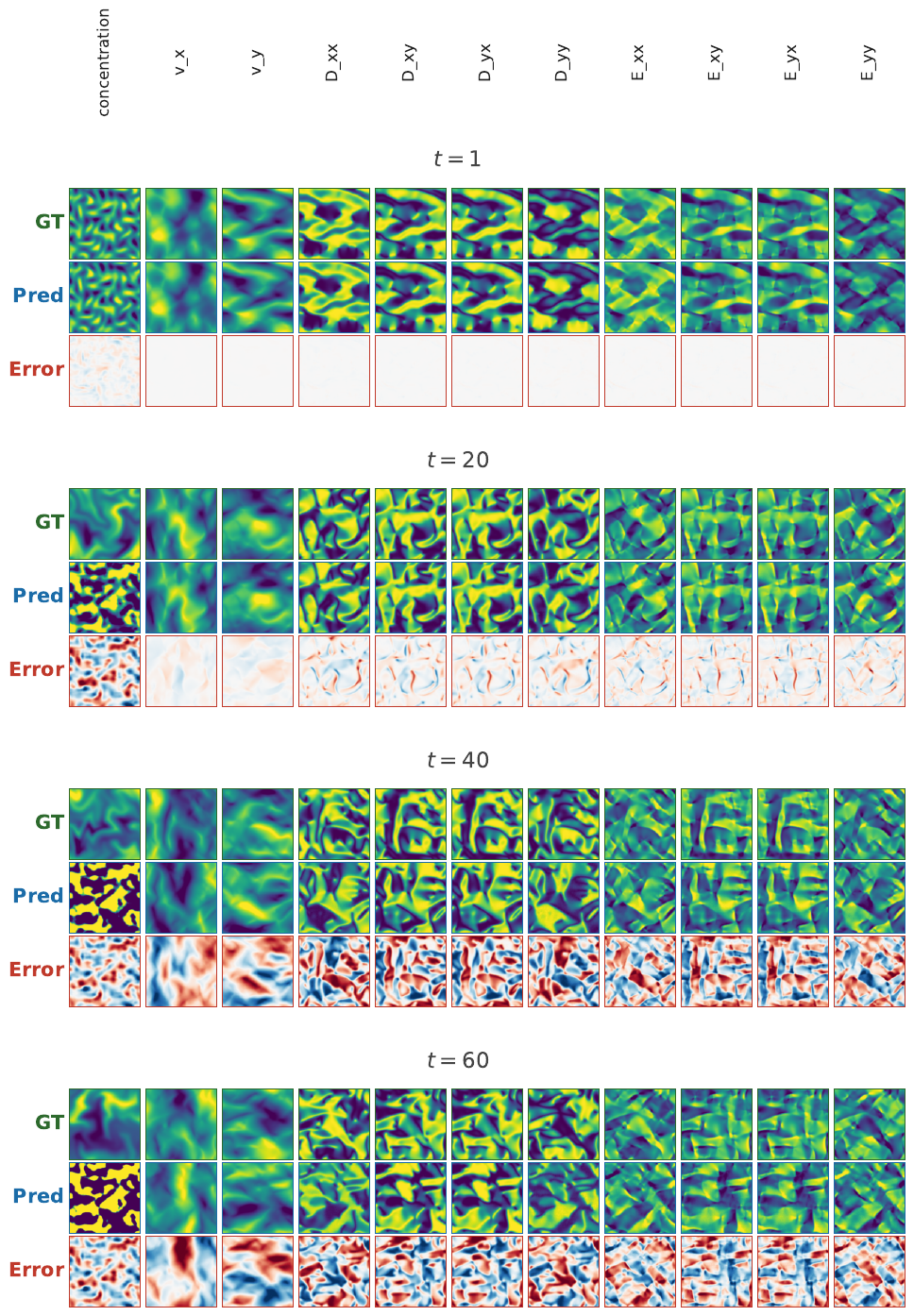}\\[4pt]
  \includegraphics[width=\textwidth,height=0.48\textheight,keepaspectratio]{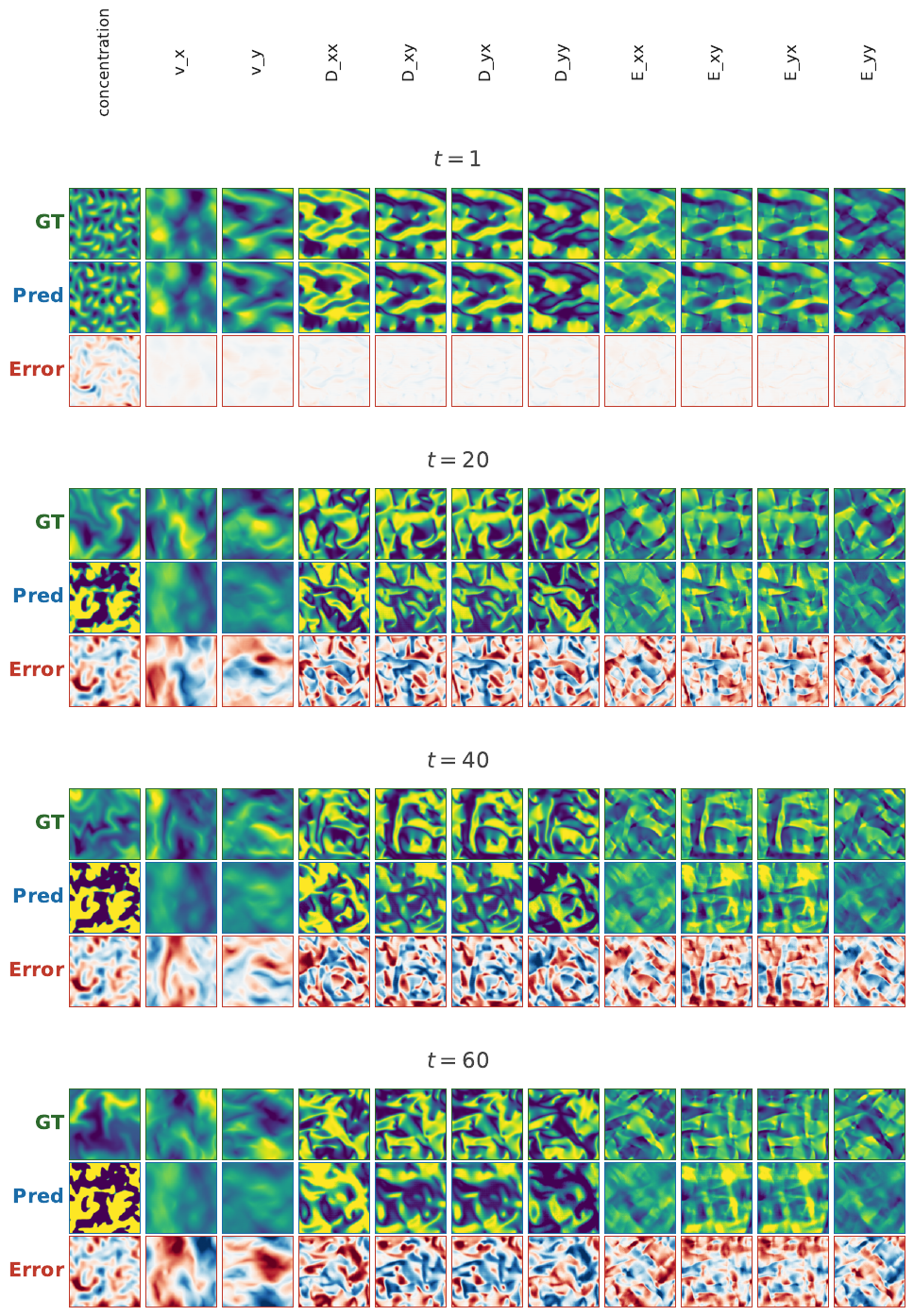}
  \caption{\textbf{Active Matter.}
    11 fields; time groups stacked vertically.
    \emph{Top}: pretrained. \emph{Bottom}: Phase~I step~20.}
  \label{fig:appendix_well_am}
\end{figure}

\begin{figure}[H]
  \centering
  \includegraphics[width=\textwidth,height=0.48\textheight,keepaspectratio]{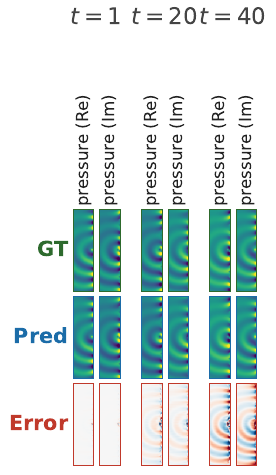}\\[4pt]
  \includegraphics[width=\textwidth,height=0.48\textheight,keepaspectratio]{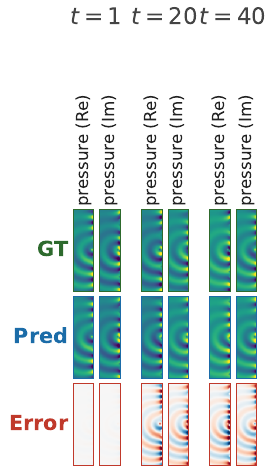}
  \caption{\textbf{Helmholtz Staircase.}
    \emph{Top}: pretrained. \emph{Bottom}: Phase~I step~20.}
  \label{fig:appendix_well_helm}
\end{figure}

\begin{figure}[H]
  \centering
  \includegraphics[width=\textwidth,height=0.48\textheight,keepaspectratio]{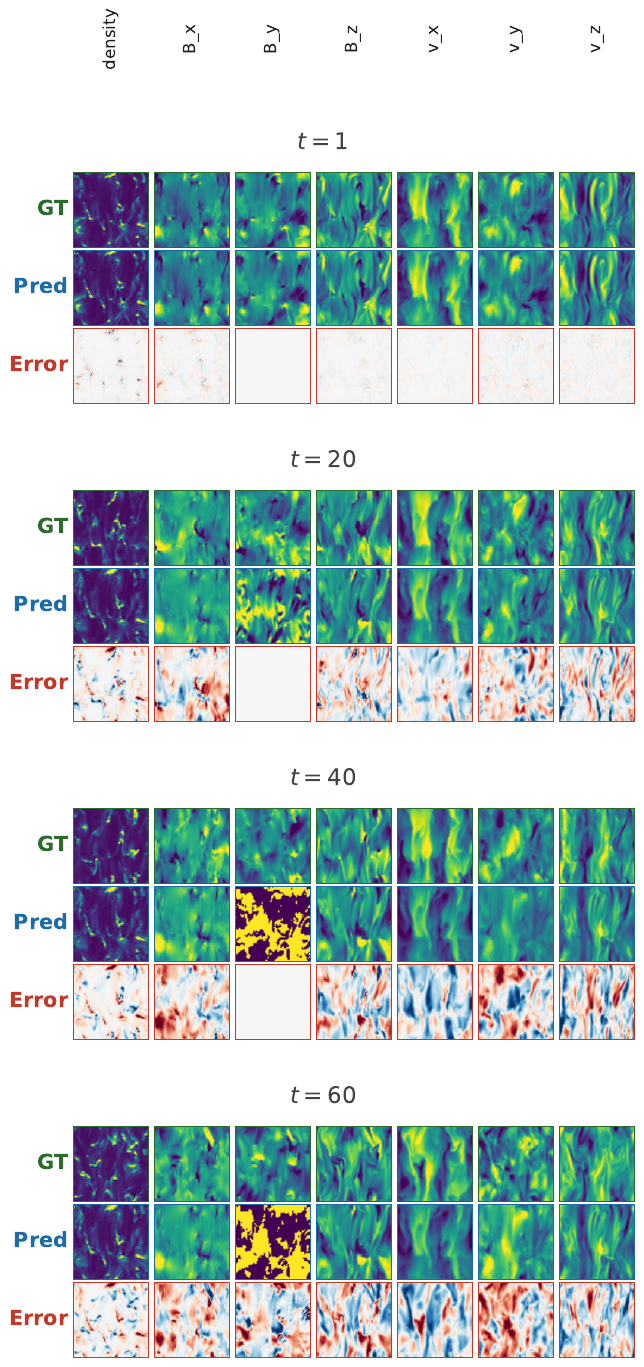}\\[4pt]
  \includegraphics[width=\textwidth,height=0.48\textheight,keepaspectratio]{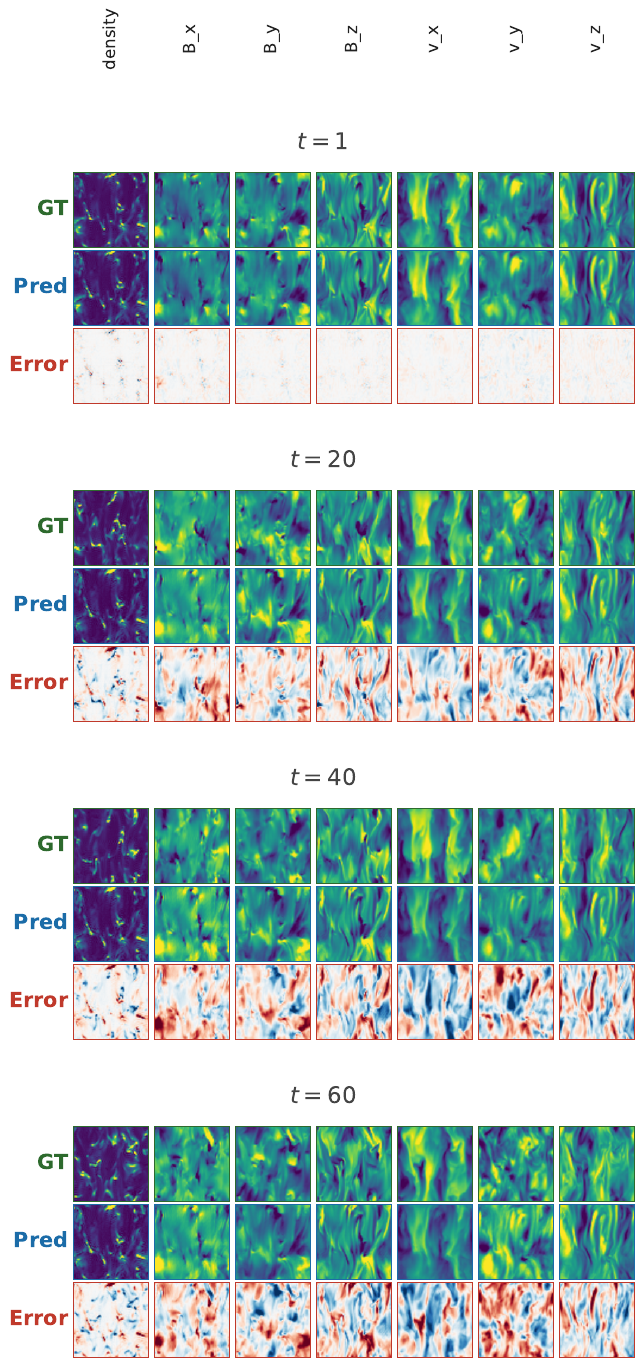}
  \caption{\textbf{MHD-64.}
    7 fields from a central $z$-slice of the 3D volume; time groups stacked vertically.
    \emph{Top}: pretrained. \emph{Bottom}: Phase~I step~20.}
  \label{fig:appendix_well_mhd}
\end{figure}

\begin{figure}[H]
  \centering
  \includegraphics[width=\textwidth,height=0.48\textheight,keepaspectratio]{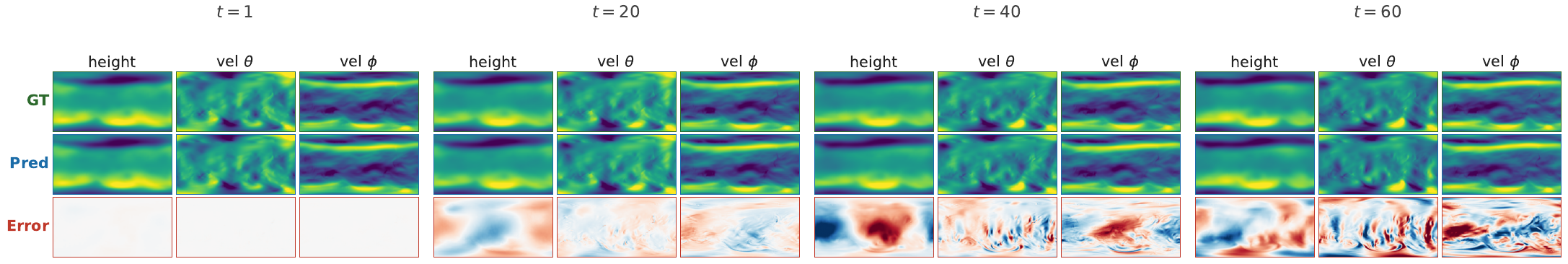}\\[4pt]
  \includegraphics[width=\textwidth,height=0.48\textheight,keepaspectratio]{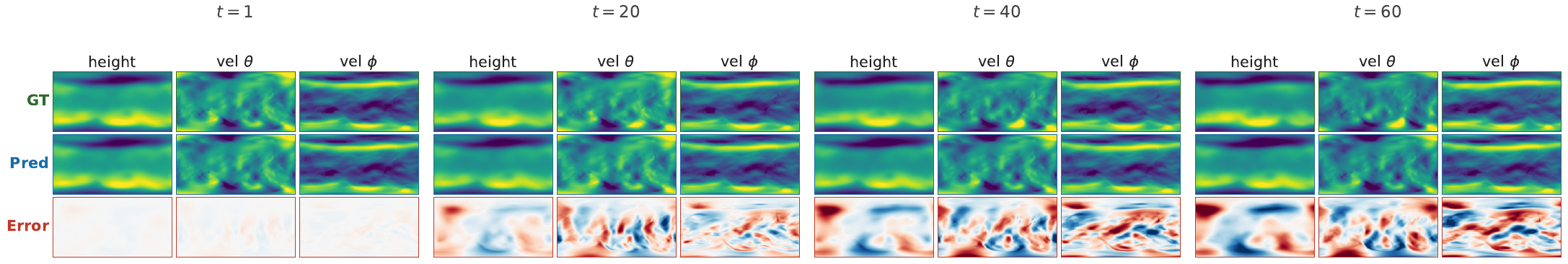}
  \caption{\textbf{PlanetSWE.}
    \emph{Top}: pretrained. \emph{Bottom}: Phase~I step~20.}
  \label{fig:appendix_well_pswe}
\end{figure}

\begin{figure}[H]
  \centering
  \includegraphics[width=\textwidth,height=0.48\textheight,keepaspectratio]{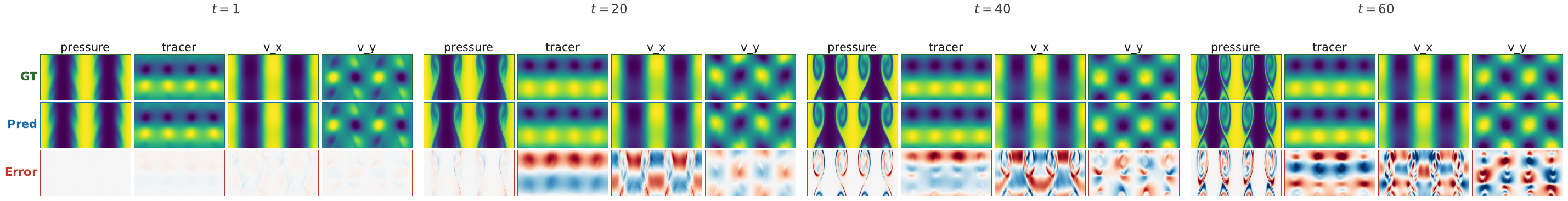}\\[4pt]
  \includegraphics[width=\textwidth,height=0.48\textheight,keepaspectratio]{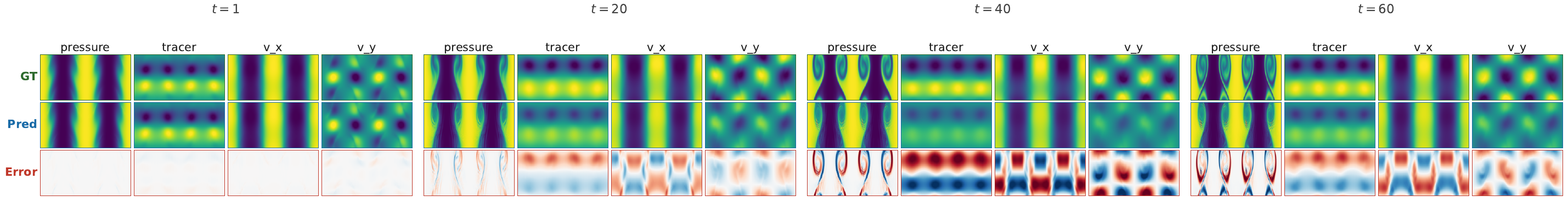}
  \caption{\textbf{Shear Flow.}
    \emph{Top}: pretrained. \emph{Bottom}: Phase~I step~20.}
  \label{fig:appendix_well_sf}
\end{figure}

\begin{figure}[H]
  \centering
  \includegraphics[width=\textwidth,height=0.48\textheight,keepaspectratio]{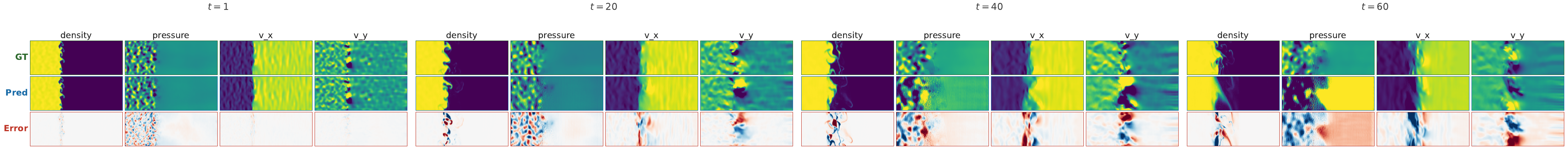}\\[4pt]
  \includegraphics[width=\textwidth,height=0.48\textheight,keepaspectratio]{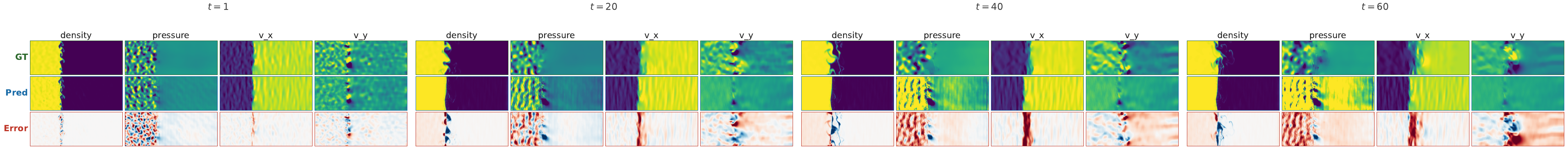}
  \caption{\textbf{Turbulent Radiative Layer 2D.}
    \emph{Top}: pretrained. \emph{Bottom}: Phase~I step~20.}
  \label{fig:appendix_well_trl2d}
\end{figure}

\begin{figure}[H]
  \centering
  \includegraphics[width=\textwidth,height=0.48\textheight,keepaspectratio]{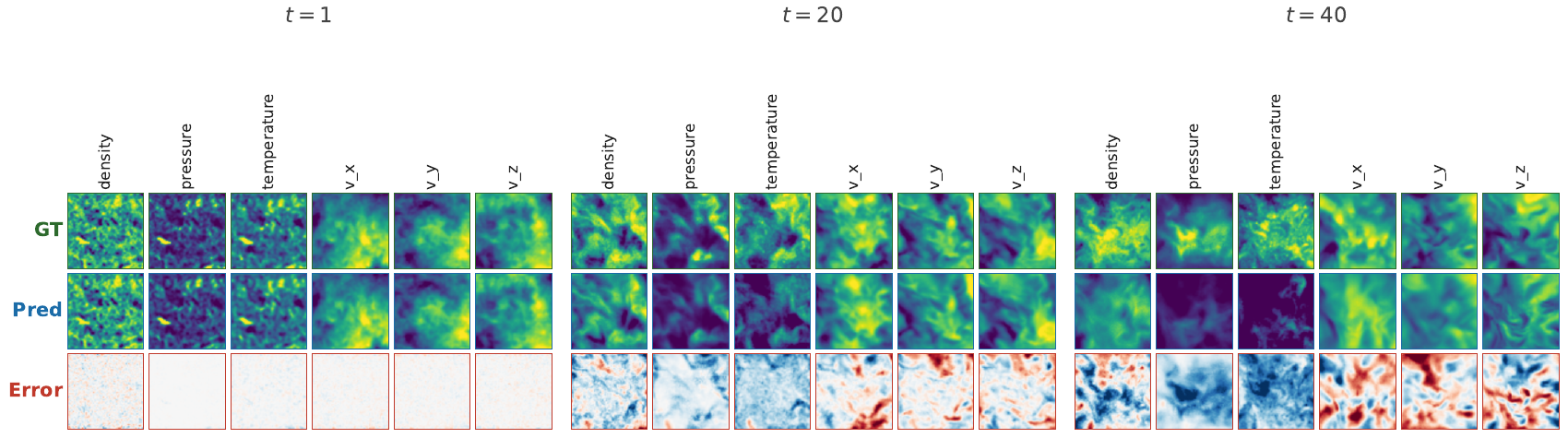}\\[4pt]
  \includegraphics[width=\textwidth,height=0.48\textheight,keepaspectratio]{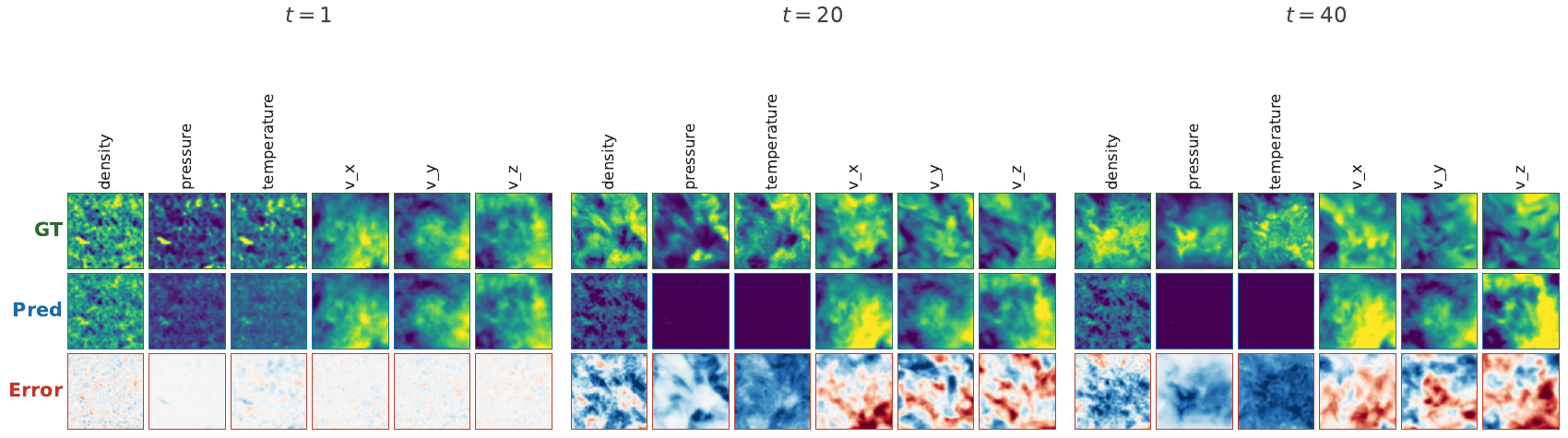}
  \caption{\textbf{Turbulence Gravity Cooling.}
    Density and temperature fields log$_{10}$-transformed before prediction.
    \emph{Top}: pretrained. \emph{Bottom}: Phase~I step~20.}
  \label{fig:appendix_well_turbcool}
\end{figure}

\newpage

\end{document}